# Ensemble Neural Networks (ENN):
# A Gradient-free Stochastic Method

Yuntian Chen[a], Haibin Chang[a], Meng Jin[a], Dongxiao Zhang[a,*]

[a] ERE and BIC-ESAT, College of Engineering, Peking University, Beijing 1000871, China
[*] Corresponding author.
E-mail address: dxz@pku.edu.cn (D. Zhang); chenyt@pku.edu.cn (Y. Chen).

**Abstract:** In this study, an efficient stochastic gradient-free method, the ensemble neural networks (ENN), is developed. In the ENN, the optimization process relies on covariance matrices rather than derivatives. The covariance matrices are calculated by the ensemble randomized maximum likelihood algorithm (EnRML), which is an inverse modeling method. The ENN is able to simultaneously provide estimations and perform uncertainty quantification since it is built under the Bayesian framework. The ENN is also robust to small training data size because the ensemble of stochastic realizations essentially enlarges the training dataset. This constitutes a desirable characteristic, especially for real-world engineering applications. In addition, the ENN does not require the calculation of gradients, which enables the use of complicated neuron models and loss functions in neural networks. We experimentally demonstrate benefits of the proposed model, in particular showing that the ENN performs much better than the traditional Bayesian neural networks (BNN). The EnRML in ENN is a substitution of gradient-based optimization algorithms, which means that it can be directly combined with the feed-forward process in other existing (deep) neural networks, such as convolutional neural networks (CNN) and recurrent neural networks (RNN), broadening future applications of the ENN.

1. Introduction

Artificial neural networks (ANN) are computing systems inspired by biological neural networks that constitute animal brains. ANN is capable of approximating nonlinear functional relationships between input and output variables (Kim et al., 2018). From a mathematical perspective, a neural network can model any function up to any given precision with a sufficiently large number of basis functions (Cybenko, 1989; Hornik, 1991). In addition, we can even use much smaller models by constructing hierarchy neural networks (Delalleau & Bengio, 2011; Gal, 2016). The basic processing elements of neural networks are neurons. A collection of neurons is referred to as a layer, and the collection of interconnected layers forms the neural networks (Kim et al., 2018). A four-layer neural network is illustrated in Fig. 1 as an example. In a neuron, the output is calculated by a nonlinear function of the sum of its inputs. The connections between different neurons from adjacent layers are represented by the weights in a model. The weights adjust as learning proceeds, and they represent the strength of the signal at a connection. The nonlinear function is also called the activation function, and the most popular choices are sigmoid, tansig, and ReLU (Li et al., 2015).

ANN has been widely applied to solving real-world engineering problems, and the following three topics are significant for effective applications.

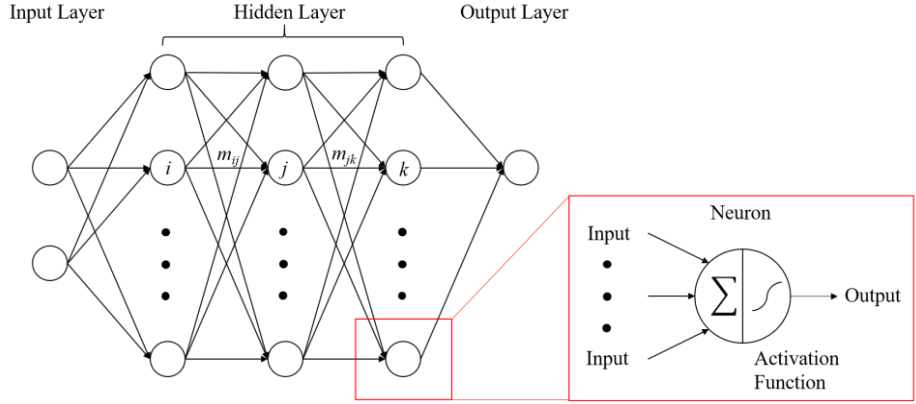

**Fig. 1.** The structure of an artificial neural network. $m_{ij}$ denotes the weight between the i[th] neuron in a layer and the j[th] neuron in the next layer. A neuron is a combination of a linear summation of inputs and an activation function.

The first topic is uncertainty quantification. Uncertainty is inevitable in all kinds of prediction models, including neural networks. Predictive uncertainty results from data uncertainty caused by noisy data, and model uncertainty comes from model parameters and model structure. Uncertainty quantification determines how much confidence one has in a certain prediction. This information is desirable in numerous fields that have the possibility to directly or indirectly affect human life, and control of them has been gradually handed-over to automated systems (Gal, 2016), such as life sciences (Herzog & Ostwald, 2013; Acharya et al., 2018) and autonomous vehicles (Widrow et al., 1994; Tian et al., 2018).

The second topic concerns data availability. Although data are the most precious resource in machine learning, data collection is very expensive and time-consuming in many real-world engineering problems. For example, in the field of gas resource evaluation in petroleum engineering, adsorbed gas content estimation is significant (Wu et al., 2014). However, an adsorption experiment could take a week to collect a single pair of data, and it is normal to spend millions of dollars in coring processes to obtain experimental material. Thus, most adsorbed shale gas datasets comprise less than 100 data (Chen et al., 2017), which hinders the application of neural networks. Data availability is especially important for deep learning, in which tens of thousands of weights need to be trained (He et al., 2016).

The final topic is not yet critical, but has the potential to greatly broaden the scope of neural network applications. In some circumstances, it is desirable to have a gradient-free optimization method. For example, in the field of brain-inspired computing, the Hodgkin-Huxley (HH) model (Hodgkin & Huxley, 1952) is utilized as the neuron model rather than the traditional McCulloch-Pitts (MCP) model (McCulloch & Pitts, 1943), in which the neuron structure is a linear combination of inputs with an activation function. Although the HH model is much more elaborate and biomimetic, and thus more accurate, it is described by a set of nonlinear differential equations and obtaining the derivatives is challenging. A gradient-free optimization method could be applied to natural language processing, as well. Bilingual evaluation understudy is an algorithm for evaluating the quality of text that has been machine-translated (Papineni et al., 2002; Reiter, 2018). However, it is difficult to build a loss function based on this evaluation criterion since it is not differentiable. However, this will no longer pose a problem if we can find a gradient-free optimization method.

Considering the aforementioned problems, a salient question is: are there any alternatives for the optimization method in a neural network that are able to perform uncertainty analysis and perform well with a small dataset, but do not rely on derivative calculations?

These obstacles are encountered in numerous engineering fields, such as petroleum engineering. Uncertainty quantification is critical because underground geological parameters are highly heterogeneous. High-dimension models are always solved based on a small dataset due to the expensive and time-consuming data collection. It is also difficult to identify gradients of a target variable with respect to model parameters because the corresponding physical models are highly nonlinear and too complicated to solve analytically. In response to these problems, the ensemble randomized maximum likelihood algorithm (EnRML) is proposed by Gu and Oliver (2007) in the field of history matching in petroleum engineering. History matching is an inverse modeling method, which adjusts a model of a reservoir until it closely reproduces its past behavior (Oliver et al., 2008; Stordal & Nævdal, 2018). It should be mentioned that the word "ensemble" here indicates a different meaning from that in ensemble averaging (Naftaly et al., 1997). In the former, it means the ensemble of realizations generated from the same model, rather than multiple models in the latter. The most prominent feature of the EnRML is that it constitutes a gradient-free optimization method because covariance matrices computed from the realizations are utilized for optimization instead of search gradients. Moreover, the EnRML is designed to solve high-dimensional problems, which is an advantage over other gradient-free methods, such as the covariance matrix adaptation evolution strategy (CMA-ES) (Hansen & Kern, 2004). Chen and Oliver (2010) have successfully solved a 267300-dimensional SPE benchmark problem with the EnRML based on 104 realizations.

The objective of this study is to find a method that is capable to perform uncertainty analysis, perform well with a small dataset, and does not rely on derivative calculations. To achieve this objective, ensemble neural networks (ENN) is proposed based on the EnRML algorithm. In the ENN, the feed-forward process is the same as the common fully-connected neural networks, but the network training process is adjusted by substituting the EnRML for the traditional gradient descent algorithm. Uncertainty quantification is straightforward in the ENN since it is based on the Bayesian theorem. In addition, the ENN is not sensitive to data size, and its optimization process does not necessitate the calculation of derivatives.

The ENN is a gradient-free stochastic method, which combines the EnRML method of historical matching with neural networks for the first time. The ENN method also shows that the neural networks can be trained through correlation information from stochastic realizations without calculating the derivatives. This study verifies the characteristics of the ENN through three computational experiments, which are regression based on a toy dataset, sanity check based on a highly nonlinear ideal dataset, and generalization test based on real-world datasets.

## 2. Methodology

*2.1 Bayesian framework*

To analyze model uncertainty, we build our model based on the Bayesian theorem and solve the problem from a probabilistic perspective. The goal is to optimize the model parameters by maximizing the posterior probability of model parameters given the training dataset. In this work, the model refers to any type of neural network, and the model parameters are the weighting coefficients therein. The posterior is defined as:

$$p(m|d_{obs}) = \frac{p(m)p(d_{obs}|m)}{p(d_{obs})} \propto p(m)p(d_{obs}|m) \tag{1}$$

where $m$ denotes the model parameters; $d_{obs}$ denotes the observed data; $p(m|d_{obs})$ denotes the posterior probability; $p(m)$ denotes the prior probability of $m$; and $p(d_{obs}|m)$ denotes the likelihood function.

Assuming that the observation is equivalent to the sum of the estimation results and normally-distributed stochastic errors, the prior probability and the likelihood function are given in Appendix A. The posterior probability distribution can be obtained by multiplying the prior probability and the likelihood function, which is shown as (Oliver et al., 2008):

$$p(m|d_{obs}) \propto exp\left[-\frac{1}{2}(m-m_{pr})^{\mathrm{T}}C_{\mathrm{M}}^{-1}(m-m_{pr}) - \frac{1}{2}\left(g(m)-d_{obs}\right)^{\mathrm{T}}C_{\mathrm{D}}^{-1}\left(g(m)-d_{obs}\right)\right] \tag{2}$$

where $m_{pr}$ denotes the prior estimate of the model parameters; $C_{\mathrm{M}}$ denotes the covariance of the prior model parameters; $g(m)$ is a function that maps $m$ to an estimation value; and $C_{\mathrm{D}}$ denotes the covariance matrix of the normally-distributed stochastic errors.

The objective function $O(m)$ is defined as proportional to the posterior probability of the model parameters, which is given as (Chang et al., 2017):

$$O(m) = \frac{1}{2}(m-m_{pr})^{\mathrm{T}}C_{\mathrm{M}}^{-1}(m-m_{pr}) + \frac{1}{2}\left(g(m)-d_{obs}\right)^{\mathrm{T}}C_{\mathrm{D}}^{-1}\left(g(m)-d_{obs}\right) \tag{3}$$

Under this framework, the optimal model parameters are those that maximize the posterior probability in Eq. (2), which equals to minimizing the objective function in Eq. (3). Intuitively, the first term in the objective function is proportional to the square of the difference between the trainable parameter $m$ and the prior estimate $m_{pr}$, which denotes the model mismatch. The second term is calculated based on the difference between the estimation value and the observation, which denotes the data mismatch. The model mismatch term not only helps the model avoid overfitting, which is similar to the regularization term in a common loss function, but also utilizes the prior information on the model parameters. The covariance matrixes $C_M$ and $C_D$ are used to nondimensionalize the model mismatch and data mismatch so that they are on the same scale.

*2.2 Ensemble randomized maximum likelihood algorithm (EnRML)*

The purpose of the EnRML is identifying a set of model parameters that maximize the posterior probability, which is equivalent to minimizing the objective function. Using the Gauss-Newton method, the iterative update formula takes the following form (Bertsekas, 1999; Chen & Oliver, 2013):

$$m^{l+1} = m^l - H(m^l)^{-1} \nabla O(m^l)$$
$$= m^l - \left[(1+\lambda_l)C_M^{-1} + G_l^T C_D^{-1} G_l\right]^{-1} \left[C_M^{-1}(m^l - m_{pr}) + G_l^T C_D^{-1}(g(m^l) - d_{obs})\right] \quad (4)$$

where $l$ denotes the iteration index; $H(m^l)$ denotes the modified Gauss-Newton Hessian matrix with the form $\left[(1+\lambda_l)C_M^{-1} + G_l^T C_D^{-1} G_l\right]$; and $G_l$ denotes the sensitivity matrix (i.e., gradient) taking a value at $m^l$. In early iterations, the data mismatch is always large, which probably results in over-update and bad iteration performance. Thus, a multiplier $\lambda$ is applied to $C_M$ in the Hessian term to mitigate the influence of large data mismatch (Li et al., 2003). The method for determining $\lambda$ will be introduced in Appendix B.

The calculation of Eq. (4) requires the inverse of the matrix $((1+\lambda_l)C_M^{-1} + G_l^T C_D^{-1} G_l)$ with size $N_m \times N_m$, where $N_m$ is the number of model parameters. For both problems of small data and those casted as online or mini-batch learning, $N_m$ is larger than the number of data points ($N_d$). Thus, a method with an inverse of a $N_d \times N_d$ matrix problem is preferred.

To reduce computation complexity, when $N_m > N_d$ the following two equalities are used to reformulate Eq. (4), the theoretical derivation of which is provided in Appendix C (Golub & Van Loan, 2012):

$$(C_M^{-1} + G_l^T C_D^{-1} G_l)^{-1} = C_M - C_M G_l^T (C_D + G_l C_M G_l^T)^{-1} G_l C_M \quad (5)$$

$$(C_M^{-1} + G_l^T C_D^{-1} G_l)^{-1} G_l^T C_D^{-1} = C_M G_l^T (C_D + G_l C_M G_l^T)^{-1} \quad (6)$$

Using Eq. (5) and Eq. (6), Eq. (4) can be rewritten as:

$$m^{l+1} = m^l - \frac{1}{1+\lambda_l}[C_M - C_M G_l^T((1+\lambda_l)C_D + G_l C_M G_l^T)^{-1} G_l C_M] C_M^{-1}(m^l - m_{pr})$$
$$- C_M G_l^T((1+\lambda_l)C_D + G_l C_M G_l^T)^{-1}(g(m^l) - d_{obs}) \tag{7}$$

where we need to calculate the inverse of a $N_d \times N_d$ matrix $((1+\lambda_l)C_D + G_l C_M G_l^T)^{-1}$.

Furthermore, a group of realizations of the model parameters are updated, so that the model uncertainty can be obtained based on the ensemble of realizations. Let $j$ denote the realization index, and Eq. (7) can be rewritten for different realizations (Oliver et al., 2008):

$$m_j^{l+1} = m_j^l - \frac{1}{1+\lambda_l}[C_{M_l} - C_{M_l}\bar{G}_l^T((1+\lambda_l)C_D + \bar{G}_l C_{M_l}\bar{G}_l^T)^{-1}\bar{G}_l C_{M_l}]C_M^{-1}(m_j^l - m_{pr,j})$$
$$- C_{M_l}\bar{G}_l^T((1+\lambda_l)C_D + \bar{G}_l C_{M_l}\bar{G}_l^T)^{-1}(g(m_j^l) - d_{obs,j}) \quad j=1,...,N_e \tag{8}$$

where $C_{M_l}$ in the Hessian matrix term is the covariance matrix of the updated model parameters at the $l^{th}$ iteration step; $C_M$ outside of the Hessian matrix term is the prior model variable covariance, which does not change with iterations; $\bar{G}_l$ denotes the average sensitivity matrix (i.e., gradient) taking a value at $\bar{m}_l$, which is the average of model parameters; $d_{obs,j}$ is a perturbed observation sampled from a multivariate Gaussian distribution with mean $d_{obs}$ and covariance $C_D$; and $N_e$ represents the number of realizations.

Eq. (8) is an effective formula to optimize the model parameters and determine the model uncertainty. However, it is a gradient-based optimization method. We can adopt the following approximations to substitute covariance and cross-covariance for the gradients (Zhang, 2001; Reynolds et al., 2006):

$$C_{M_l,D_l} \approx C_{M_l}\bar{G}^T$$
$$C_{D_l} \approx \bar{G}C_{M_l}\bar{G}^T \tag{9}$$

where $C_{M_l,D_l}$ denotes the cross-covariance between model parameters $m$ and estimation values $g(m)$; and $C_{D_l}$ denotes the covariance of estimation values $g(m)$. The proof of these two approximations is provided in Appendix D.

Using Eq. (9), we can rewrite Eq. (8) as:

$$\begin{aligned} m_j^{l+1} = m_j^l &- \frac{1}{1+\lambda_l}\left[C_{M_l} - C_{M_l,D_l}\left((1+\lambda_l)C_D + C_{D_l}\right)^{-1}C_{M_l,D_l}^{T}\right]C_M^{-1}(m_j^l - m_{pr,j}) \\ &- C_{M_l,D_l}\left((1+\lambda_l)C_D + C_{D_l}\right)^{-1}\left(g(m_j^l) - d_{obs,j}\right) \quad j=1,\ldots,N_e \end{aligned} \quad (10)$$

where $C_{M_l,D_l}$ denotes the cross-covariance between the updated model parameters and the prediction at iteration step $l$ based on the ensemble of realizations; and $C_{D_l}$ is the covariance of the predicted data.

---

**Algorithm 1** Minimize the objective function in the EnRML

**Input:** $x$ and $y$

**Trainable parameter:** $m$

**Hyper-parameters:** $m_{pr}$, $C_D$, and $C_M$ (determined based on prior information)

For $j = 1,\ldots,N_e$

1. Generate realizations of measurement error $\varepsilon$ based on its probability distribution function (PDF);

2. Generate initial realizations of the model parameters $m_j$ based on prior PDF;

3. Calculate the observed data $d_{obs}$ by adding the measurement error $\varepsilon$ to the target value $y$;

**repeat**

    **Step 1:** Compute the predicted data $g(m_j)$ for each realization based on the model parameters;

    **Step 2:** Update the model parameters $m_j$ according to Eq. (10). The $C_{M_l,D_l}$ and $C_{D_l}$ are calculated among the ensemble of realizations. Thus, the ensemble of realizations is updated simultaneously;

**until** the training loss has converged.

---

In the EnRML, to construct a group of realizations, the randomized maximum likelihood algorithm (RML) is employed to generate the samples in each realization. The observed data utilized in each realization are the sum of the target value and a stochastic measurement error, and the initial model parameters are generated by sampling the prior probability distribution function for model parameters. The EnRML can be summarized as Algorithm 1. The hyper-parameters in EnRML are all determined based on prior information. It can be seen from Eq. (10) that the time complexity of this algorithm is $O(d^3 + m^2d + md^2)$, where $m$ denotes the number of model parameters and $d$ denotes the number of the training data.

In the EnRML, the calculation of derivatives is not required in the optimization process, and most variables in Eq. (10) are easily accessible statistics. Moreover, the realizations are related to each other, and the iteration of each realization utilizes information from the whole ensemble.

*2.3 Ensemble neural networks (ENN)*

In the ENN, the EnRML is combined with the feed-forward neural network by taking the weights in the neural network as the model parameters $m$ in the EnRML. The inputs from the training data are taken as a fixed part of the feed-forward process in the ENN. The essence of the ENN is illustrated in Fig. 2. Firstly, a group of $N_e$ realizations of weights are generated for a feed-forward neural network of a given architecture. It is noted that the same methodology can be extended to account for model uncertainties for which the architecture varies among realizations. Then, the same set of input variables are used to calculate the prediction values of each realization based on the neural network architecture and the corresponding weights in each realization. Finally, the EnRML is applied to the optimization process, and the weights are updated according to Eq. (10). An end-to-end example is provided in the Supplementary Material to demonstrate the calculation process of the ENN.

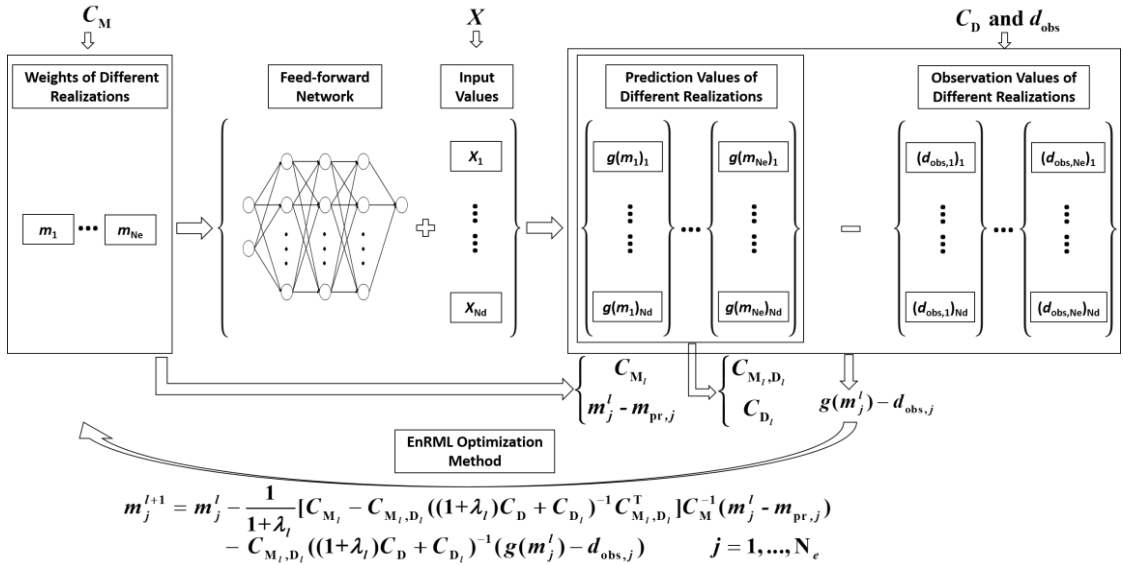

**Fig. 2.** Flow chart of ensemble neural networks. The EnRML algorithm is used as an optimization method in the ENN. An example of the calculation process is provided in the Supplementary Material.

The ENN is able to perform uncertainty quantification since it is calculated from a probabilistic perspective. In fact, the usage of the ensemble of realizations receives a double advantage in the ENN. It not only provides essential information about covariances to optimize the weights, but also enables the model to solve small data problems by generating artificial training data without additional effort. It should be mentioned that, in the ENN, the model $g(m_j)$ can be the feed-forward process of convolutional neural networks (CNN), recurrent neural networks (RNN), or other neural networks. In this study, the ENN is casted in a batch learning mode. However, the ENN can be conveniently formulated as sequential (online) learning. Moreover, the embedded EnRML is embarrassingly parallel.

The ENN seems to be similar to other uncertainty quantification methods, such as Bayesian neural networks (BNN) and Monte Carlo neural networks (MCNN) (MacKay, 1992; Gal, 2016). Nevertheless, the minimization method, determination of hyperparameters, and methods for solving the Hessian matrix are totally different for the ENN and the BNN. Regarding the MCNN, each realization has an independent forward process and an independent backward process to the other realizations. In contrast, realizations in the ENN share information for simplifying the update in the backward process. The forward process in the ENN can not only obtain estimation values, but also obtain the shared covariance matrices. Detailed comparisons of ENN and other similar models are provided in Appendix E.

## 3. Experiments

To test the performance of the ENN, several computational experiments are carried out in this section. The main purposes of the experiments are: 1) to comprehensively understand the ENN method, such as estimation uncertainty; 2) to analyze the capability of the ENN to generate accurate estimations, especially when different scales of measurement errors exist in observations and different sizes of training datasets are applied; and 3) to compare the performance of the ENN with traditional BNN methods.

In the following subsections, we introduce three different experiments. The descriptions and purposes of the experiments are summarized in Table 1. In the first experiment, a one-dimensional toy regression dataset is utilized. This dataset was proposed by Hernandez-Lobato and Adams (2015) to evaluate models' capability to perform uncertainty quantification. In the second experiment, a sanity check is performed to test whether the ENN method is functional. If the ENN does not work on an ideal problem, it is impossible for the ENN to solve real engineering problems. Moreover, the ENN is compared with the BNN to analyze their performance under different scenarios. In the last experiment, four real-world datasets are utilized as benchmarks to compare the generalization capability between the ENN and the BNN. Three of the datasets are from the UCI Repository of Machine Learning Databases (Lichman, 2013), and the final one is a highly nonlinear problem in petroleum engineering.

**Table 1**

Descriptions and purposes of the computational experiments.

| Experiment | Purpose |
| --- | --- |
| Regression based on a one-dimensional toy dataset | Evaluate models' capability to perform uncertainty quantification. |
| Sanity check based on a highly nonlinear ideal dataset | Confirm the feasibility of the ENN; Evaluate the performance and convergence process of the ENN; Analyze the influence of network architecture, training data size and scale of observation errors on the ENN, and compare its performance with the BNN. |
| Generalization test based on real-world datasets | Evaluate the generalization capability of the ENN by comparing with the BNN. |

## 3.1 Experiment on a toy dataset

Hernandez-Lobato and Adams (2015) proposed a toy dataset to evaluate models' capability to perform uncertainty quantification, which is widely used by other researchers (Lakshminarayanan et al., 2017; Louizos & Welling, 2017). This toy dataset consists of 20 training examples drawn as $y = x^3 + \varepsilon$, where $\varepsilon$ follows a normal distribution with mean 0 and variance 9, and $x \in (-4, 4)$. To illustrate the influence of various observations on estimation (learning) accuracy and associated uncertainty, we drew different sizes of training samples from different intervals in this study. We used the same neural network architecture as Hernandez-Lobato and Adams (2015), and ReLU was taken as an activation function. The results are shown in Fig. 3. For the case of no observations, as shown in Fig. 3(a), the estimated function is a flat line, and the estimation uncertainty results from the prior distribution of ENN weights. In Fig. 3(b), when 10 training samples are randomly drawn from (-2, 2), the estimated function is still a flat line, and the corresponding uncertainty is almost unchanged because the training area is too small to describe the cubic function. It is shown from Fig. 3(c) to Fig. 3(f) that estimation and uncertainty are improved with the increase of training samples, which indicates that this method is particularly suitable for active learning. The ENN is capable of making reasonable estimations even at points far from the training data, although the cubic function increases rapidly there. It is noted that the samples in Fig. 3(d) are drawn from the same interval as the experiment in a previous study (Hernandez-Lobato & Adams, 2015). In that study, the estimation uncertainty is so large outside of the training area that the boundaries of estimation uncertainty are even contrary to the real trend of the cubic function. However, as shown in Fig. 3(d), the estimation uncertainty of the ENN exhibits the same trend as the ground truth target function, even outside of the training area. This toy data set provides us with a good understanding of the estimation and uncertainty quantification of the ENN.

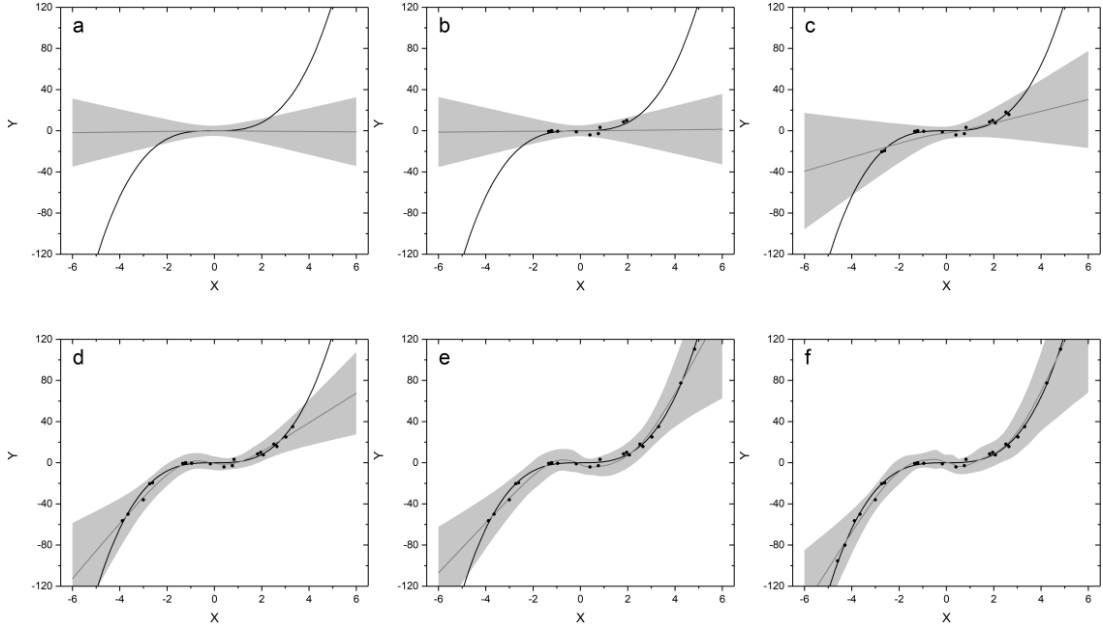

**Fig. 3.** Estimation and uncertainty quantification of the ENN on a toy dataset. The black line and gray line denote ground truth target function and estimated function, respectively. The observed noisy training examples are represented by black points, and the gray area illustrates estimation uncertainty, which corresponds to the predicted mean along with three standard deviations. Training examples of different sizes are drawn from different training areas: (a) no observations; (b) 10 training samples from (-2, 2); (c) 15 from (-3, 3); (d) 20 from (-4, 4); (e) 22 from (-4, 5); and (f) 24 from (-5, 5).

*3.2 Experiment on a highly nonlinear ideal dataset*

In the ideal regression problem, independent variables and true weights are randomly sampled from a normal distribution. The target variables are calculated based on a given neural network with a certain architecture. The goal is using the ENN to estimate the weights, given the neural network's architecture and the observations. The fact that a true solution exists in this ideal regression problem makes this experiment a sanity check. If the ENN does not work on this ideal problem, it is impossible for the ENN to solve real engineering problems. Furthermore, this experiment can be utilized to examine the influences of network architecture, training data size, and scale of observation errors on estimation accuracy. Regarding the parameters in the ideal artificial dataset, the two independent variables are sampled from a normal distribution with mean 0 and standard deviation 10. The weights are drawn from a standard normal distribution. The given neural network architecture (reference architecture) is a fully connected network with three hidden layers and one output layer. The number of neurons is four, four, 10 and one, respectively, for each layer. The artificial dataset has 70 pairs of training data and 30 observations as the testing data.

*3.2.1 Performance and convergence process of ENN*

Regarding the hyperparameters in the ENN, the mean and standard deviation of measurement errors are 0 and 0.002, respectively. The prior covariance matrix of model parameters is set to be an identity matrix. The number of realizations in the ensemble is set as 100. The aforementioned setting is denoted as default hyperparameters of the ENN in this study. Initially, the ENN architecture is set to be the same as the reference architecture to avoid the influence of network architecture. As is shown in Fig. 4(a), both training loss and test loss decrease rapidly in the first 25 iteration steps, and test loss converges to 0.177 at the $92^{nd}$ iteration. The simultaneous decrease in training loss and test loss indicates that over-fitting has not occurred, although the number of training data (70) is less than that of weights (93). The scatter plot of the observed target values versus their corresponding estimation values of the test dataset is presented in Fig. 4(b). The closer the distribution of points is to the $45°$ diagonal, the better is the estimation. This shows that the ENN performs well, even when it suffers from a lack of data. This advantage resulted from the ensemble method, in which the small training dataset is actually enlarged by the duplication with random disturbance. In essence, the ensemble method is similar to the data augmentation process in image recognition, such as horizontal and vertical flip, rotation, and adding noise.

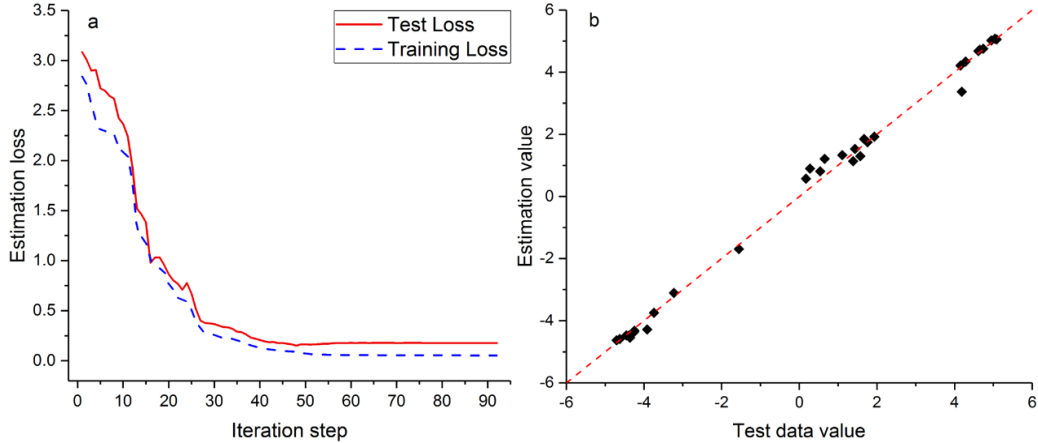

**Fig. 4.** (a) Decrease of ENN estimation loss on the test dataset and training dataset with iterations; (b) scatter plot of the observed target values versus their corresponding estimation values in the test dataset. The 45° diagonal is illustrated by the red dashed line.

In feed-forward neural networks, multiple distinct choices for the weights can give rise to the same mapping function from inputs to outputs (Bishop, 2007). For instance, if the 'tanh' activation function is applied to the network and we change the sign of all of the weights and the bias feeding into a particular hidden unit, then, for a given input pattern, the sign of the activation of the hidden unit will be reversed. This transformation can be exactly compensated by simply changing the sign of all of the weights leading out of that hidden unit. This weight-space symmetries property may cause difficulties when the Monte Carlo method is used since the realizations in Monte Carlo are independent. In other words, it is very likely to obtain totally different weights from different realizations in the Monte Carlo method, but these different weights actually have the same performance. Regarding the ENN method, although there exist many weights that are different but have a similar mapping function, the weights from different realizations may ultimately converge in a statistical sense. This convergence is resulted from the close interrelation among realizations. The convergence processes of different realizations are illustrated in Fig. 5. There are a total of 93 weights in the model, and six of them are taken as examples. The red solid line denotes the expectation value of the weights, which is equivalent to the mean of the weights from different realizations. The blue dashed lines are the upper and lower bound of the weights. The vertical distances between the blue dashed lines and the red line equal the standard deviation of the weights from different realizations. The converging processes of different weights prove that the influence of prior distribution on the final results is very small. Although the weights are broadly distributed at the beginning due to the random sampling, they converge rapidly with iteration. This means that the prior distribution does not need to be very accurate. Thus, it is reasonable to use Gaussian distribution to generate the initial realizations in ENN. To provide a more comprehensive illustration of the whole group of 93 weights, the mean and standard variance of the weights at each iteration step are illustrated by the gray scale map in Fig. 6(a) and Fig. 6(b), respectively. As is shown in Fig. 6, the means of all of the weights converge well, and the standard deviations are close to zero.

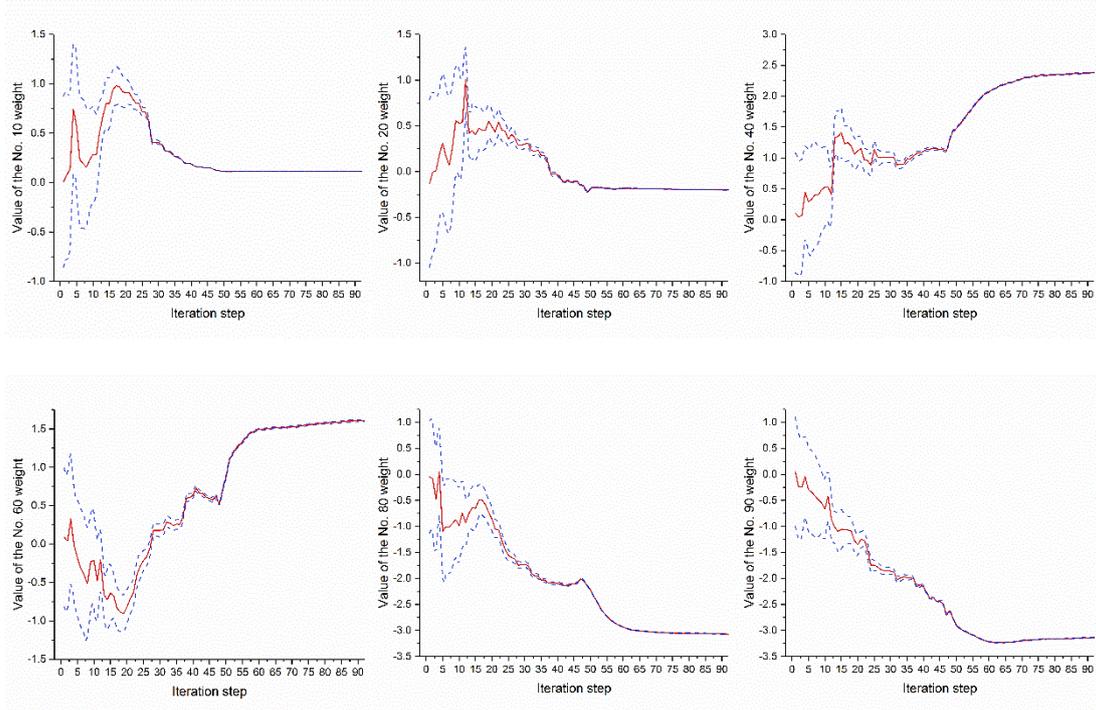

**Fig. 5.** The detailed converging process of different weights. The red solid line is the mean of different realizations. The blue dashed lines are the upper and lower bound, respectively. The gap between the red line and blue lines is equivalent to the value of the standard deviation of different weights. The weights' ID is sorted from the output layer to the input layer, which means the smaller the number, the deeper the connection.

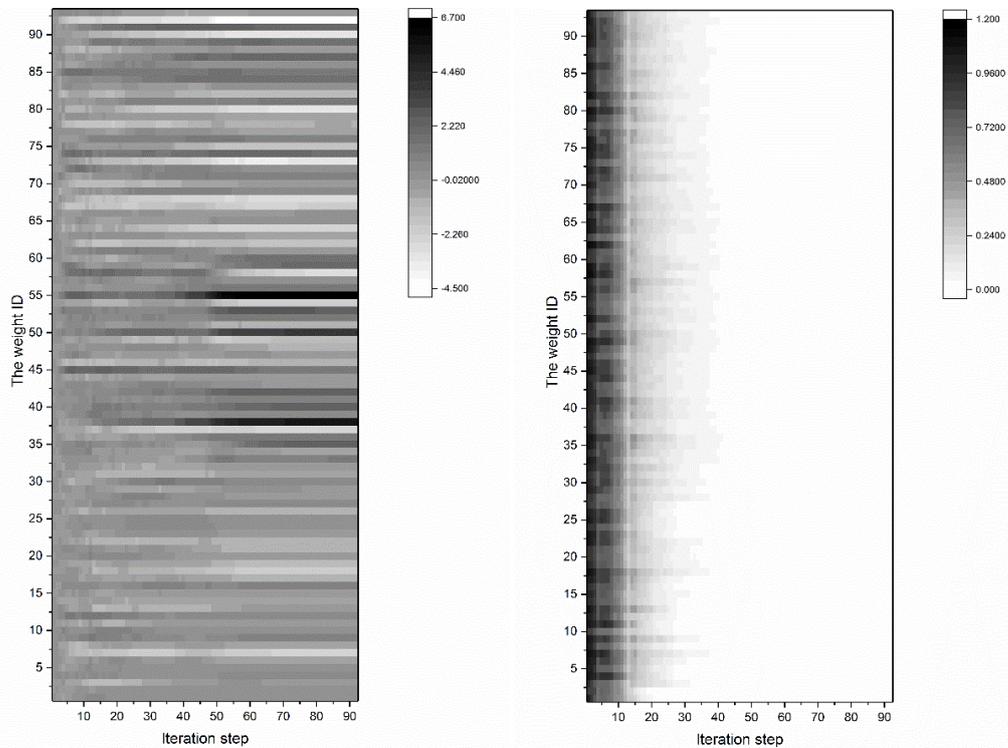

**Fig. 6.** The grey scale maps of mean and standard variance of weights: (a) The mean of weights at each iteration; (b) the standard variance of each weight at different iteration. It is illustrated that the means are converging to different values, and the standard variances are converging to 0.

*3.2.2 The influence of network architecture, training data size, and scale of observation errors*

To examine the influence of network architecture, five different neural networks with varied numbers of layers and different numbers of weights are compared with the reference neural network. All of the six neural networks have the same hyperparameters, and the only difference among them is architecture, which is shown in Table 2. The $N_t / N_w$ denotes the ratio of the number of training data to the number of weights. Architecture 2, 3, and 4 have a similar value of $N_t / N_w$ to the reference architecture, so that the comparison among these models reveals the influence of the neural networks' depth. Furthermore, architecture 1 and 5 have totally different ratios from the reference architecture, which shows the influence of the number of weights because the training data size is the same. For each architecture, estimation loss is calculated based on 50 randomly sampled independent experiments, and the results are provided as boxplots. It is shown in Fig. 7 that the neural networks with different architectures exhibit a similar performance, as long as they have a similar ratio of $N_t / N_w$. In addition, it is revealed by the comparison of architecture 1 and 5, and the reference architecture that the bigger is the ratio, the lower is the estimation loss.

**Table 2**
Architectures of different ENN models.

|  | Mean of estimation loss | Number of weights | Number of layers | Number of neurons in each layer | $N_t / N_w$ |
|---|---|---|---|---|---|
| Architecture 1 | 0.459 | 183 | 4 | 4, 10, 10, 1 | 0.4 |
| Architecture 2 | 0.269 | 91 | 3 | 6, 9, 1 | 0.8 |
| Architecture 3 | 0.247 | 91 | 4 | 4, 6, 6, 1 | 0.8 |
| Reference architecture | 0.258 | 93 | 4 | 4, 4, 10, 1 | 0.8 |
| Architecture 4 | 0.248 | 93 | 5 | 4, 4, 5, 5, 1 | 0.8 |
| Architecture 5 | 0.129 | 50 | 4 | 3, 4, 4, 1 | 1.4 |

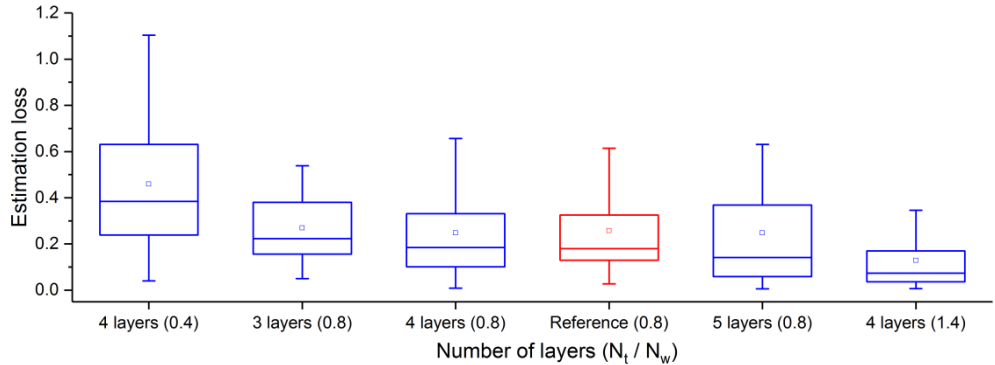

**Fig. 7.** Boxplot of estimation loss to evaluate the influence of neural network architecture. The red boxplot is the reference architecture, which is the same as the neural network to generate the ideal artificial dataset. The numbers in brackets are the ratios of the number of training data to the number of weights.

In the following part, the ENN is compared with the BNN under different sizes of training datasets and various scales of measurement errors. This series of comparisons reveals the difference of robustness between the ENN and the BNN. The average estimation losses of the ENN and the BNN with different training data sizes are illustrated in Fig. 8 by boxplot. A group of normally-distributed stochastic measurement errors with mean 0 and standard deviation 0.1 is added to the estimations. In this experiment, both the BNN and the ENN use the reference architecture. The maximum size of the training dataset is 700 and the reference architecture has 93 model parameters, which means that the maximum amount of training data is equivalent to 5.3 times the amount of model parameters. Hyperparameters in the ENN are set according to the real measurement error. It is shown in Fig. 8 that: 1) the increase of the size of the training dataset improves the performance of both the BNN and the ENN; 2) the ENN is more robust than the BNN, especially when the model suffers from a lack of data; and 3) given enough training data, the estimation accuracy of the BNN approaches that of the ENN. When the amount of training data reaches more than three times the amount of model parameters, the loss of the BNN is slightly smaller than the loss of the ENN, but both are very close to 0. This phenomenon is in line with the calculation principle of the ENN. The ENN assumes that the observation is equivalent to the sum of the estimation results and normally-distributed stochastic errors. In order to improve the model accuracy under a small training dataset, the ENN adds a normally-distributed random noise to the observation data. This approach can improve the robustness of the model when the data are insufficient, but when the data are sufficient, the artificially added noise will increase the prediction error of the model. It should be mentioned that the ENN is ideal for real-world engineering problems, in which it is difficult to obtain sufficient training data. For example, the data size of most datasets is less than 100 (Chen et al., 2017) in adsorbed shale gas problems. For these problems, the ratio of the training data size to the amount of model parameters tends to be much smaller than the maximum ratio (5.3) in Figure 8.

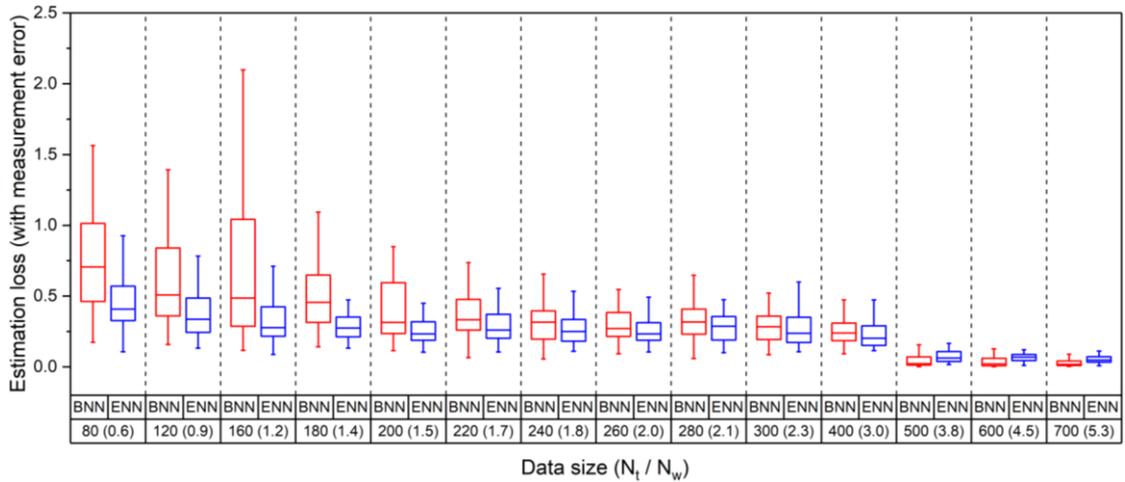

**Fig. 8.** Distribution of the estimation loss for the ENN and the BNN with the same scale of stochastic measurement errors and different data sizes. The measurement error is sampled from a normal distribution with mean 0 and standard deviation 0.1. Each boxplot is drawn based on 50 randomly sampled independent experiments. The numbers on the abscissa axis denote data size, and the numbers in brackets are the ratios of the number of training data to the number of weights.

Moreover, the influence of different scales of measurement error is examined. As shown in Eq. (A.1) in Appendix A, the observations in the training dataset are equivalent to the sum of the estimation results and normally-distributed stochastic errors (measurement errors). The measurement errors are in accordance with normal distributions and their means are zero, while their standard deviations are different. Thus, the standard deviations actually determine the scales of measurement error. In other words, the bigger is the standard deviation, the larger is the measurement error. In this experiment, the standard deviation ranges from 0.001 to 0.2, which is represented by the abscissa axis. It is shown in Fig. 9 that: 1) the decrease of the scale of measurement error improves the performance of both the BNN and the ENN since it reduces the uncertainty in the observations; and 2) the smaller interquartile range of the ENN proves the better robustness of the ENN compared to the BNN in the presence of any scale of measurement error.

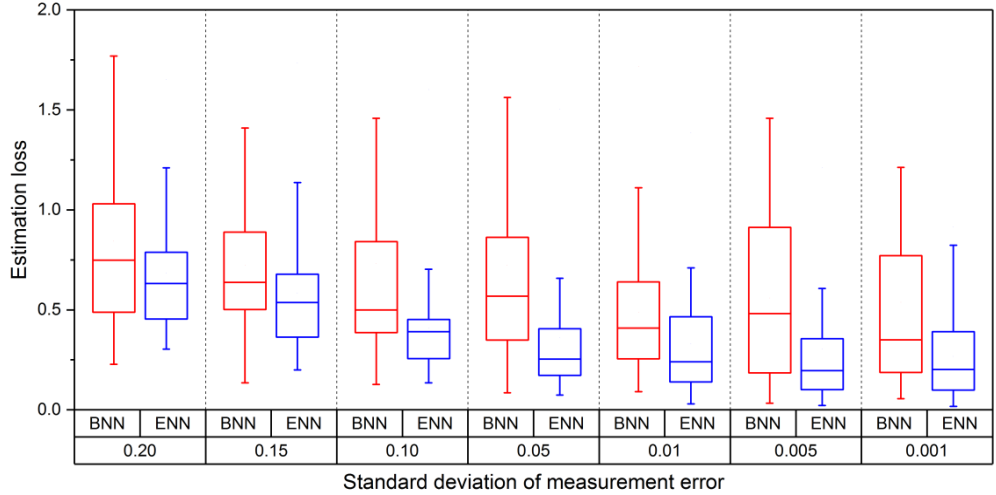

**Fig. 9.** Distribution of estimation loss for the ENN and the BNN with the same data size and different scales of stochastic measurement errors. Each boxplot is drawn based on 50 randomly sampled independent experiments. The data size is set to be 100.

*3.3 Experiment on real-world datasets*

The generalization capability of the ENN is assessed in this subsection by applying the ENN to four different datasets. The classical fully connected neural networks (FCNN) and three different kinds of BNN algorithms are examined as comparisons. The first BNN algorithm is the traditional BNN. The second one is the BNN with a validation process, which is denoted as BNN-Val. In the third method, the learning function is replaced by gradient descent with momentum (GDM). The FCNN is the most commonly used neural network, but it does not provide the probability distribution function of estimations, and uncertainty quantification cannot be performed. The BNN algorithms are chosen as the baselines since they are capable to perform uncertainty quantification. In the ENN, the aforementioned default values are utilized as hyperparameters. Regarding the training datasets, three of them are taken from the UCI Repository of Machine Learning Databases (Lichman, 2013), which are Auto MPG, Combined Cycle Power Plant, and Concrete Slump (Quinlan, 1993; Kaya et al., 2012; Yeh, 2007). Moreover, except for the datasets from the UCI database, a dataset called PRES-2D is used, which is based on a highly nonlinear water-oil two-phase flow problem. Each data point has 20 independent input variables and five output variables.

This dataset is provided and introduced in Appendix F. Regarding the network architecture, the FCNN, the three kinds of BNN methods, and the ENN method share the same three-hidden layer neural network for the same dataset. The network architecture contains four, four, and 10 hidden units for each layer for the UCI datasets. For the PRES-2D dataset, the network has 15, 10 and five hidden units for each layer since this dataset has a higher input dimension.

**Table 3**

Average estimation loss of different training methods.

| Dataset | FCNN Loss | BNN Loss | BNN-Val Loss | BNN-GDM Loss | ENN Loss |
| --- | --- | --- | --- | --- | --- |
| Auto MPG 60 | 0.164 | 0.281 | 0.271 | 0.284 | **0.146** |
| Auto MPG 80 | 0.162 | 0.263 | 0.265 | 0.260 | **0.133** |
| Auto MPG 100 | 0.150 | 0.254 | 0.255 | 0.260 | **0.122** |
| Auto MPG 150 | 0.195 | 0.271 | 0.254 | 0.265 | **0.122** |
| CCPP 60 | 0.234 | 0.419 | 0.435 | 0.411 | **0.230** |
| CCPP 80 | 0.229 | 0.423 | 0.404 | 0.427 | **0.219** |
| CCPP 100 | 0.226 | 0.383 | 0.398 | 0.370 | **0.181** |
| CCPP 150 | 0.243 | 0.368 | 0.379 | 0.333 | **0.172** |
| Concrete 60 | 0.451 | 0.428 | 0.428 | 0.427 | **0.419** |
| Concrete 80 | 0.457 | 0.392 | 0.392 | 0.392 | **0.381** |
| Concrete 100 | 0.486 | 0.430 | 0.430 | 0.417 | **0.365** |
| PRES-2D 100 | 0.117 | 0.249 | 0.249 | 0.250 | **0.109** |
| PRES-2D 150 | 0.076 | 0.224 | 0.224 | 0.224 | **0.071** |
| PRES-2D 170 | 0.068 | 0.159 | 0.165 | 0.154 | **0.066** |
| PRES-2D 180 | 0.064 | 0.085 | 0.107 | 0.106 | **0.063** |
| PRES-2D 190 | **0.058** | 0.101 | 0.074 | 0.103 | 0.063 |
| PRES-2D 200 | **0.050** | 0.061 | 0.082 | 0.061 | 0.061 |

The experiment results are shown in Table 3. The suffix of the dataset's name denotes the size of the training data. As shown in the experiment results, the performances of the methods become better with the increase of the training dataset. The average estimation losses of the BNN, the BNN-Val and the BNN-GDM are very similar, and they are all larger than the loss of the ENN, except for the last case where the losses converge. This phenomenon verifies the ENN's advantage in the case of a small training dataset. In the last two cases, the FCNN performs better than the BNN and the ENN when the data are sufficient; however, the FCNN cannot conduct uncertainty quantification, which restricts its application in real-world engineering problems. Thus, the three BNN algorithms are better baselines than the FCNN when the probability distribution functions are important. As has been discussed in section 3.2, the superior performance of the ENN resulted from the ensemble of stochastic realizations, which can avoid the over-fitting problem to some extent. In general, the validation process is a common choice for avoiding over-fitting. However, the performance of the BNN and the BNN-Val is very similar, which means that validation is not useful in small data problems. This is because the validation process reduces the size of the training dataset, which then aggravates the over-fitting problem. It is shown in PRES-2D 200 that the estimation losses of the BNN methods and the ENN become similar when there is enough data. The ENN cannot replace the commonly used FCNN for problems that have sufficient training data and do not require uncertainty quantification or gradient-free calculation. However, the ENN is a better choice when it is necessary to perform uncertain quantification and the training data are insufficient. The ENN also possesses a unique advantage for problems that require gradient-free calculations.

4. **Discussion and conclusion**

In this study, an optimization algorithm, called the EnRML, is applied to neural networks to build ensemble neural networks (ENN). The ENN is a stochastic method based on the Bayesian theorem. A group of stochastic realizations is used to obtain statistical properties, such as the covariance between weights and estimation values. In essence, the ENN method utilizes correlation information from stochastic realizations to minimize the objective function. The minimization of the objective function in the ENN is equivalent to the maximization of posterior probability. Firstly, the Gauss-Newton method is applied to the minimization of the objective function, and an iterative update formula that depends on both covariance and gradients is obtained. Secondly, the iterative update formula is reformulated to reduce computational complexity, which is desirable especially for the problem of a lack of samples. This adjustment is not necessary if the number of data is larger than the number of weights. Thirdly, the sensitivity matrix in the update formula is substituted by covariance between the updated weights and the predicted data by using first-order approximation of Taylor expansion. The finally-obtained iterative update formula does not require knowledge of derivatives, and it only depends on covariance, which can be easily calculated from the stochastic realizations.

With the ENN, it is straightforward to obtain the probability distribution functions (PDFs) of not only weights, but also estimation values from the ensemble of realizations. These PDFs could be applied to uncertainty quantification, which is beneficial for increasing neural networks' safety and accuracy. The computational experiment in section 3.1 demonstrates the capability of the ENN to perform uncertainty quantification, which is an advantage of the ENN over other classical neural networks, such as the commonly-used fully connected neural networks. Although Bayesian neural networks and Monte Carlo neural networks are able to obtain the PDFs as well, the ENN is simpler. In the backward process of the ENN, the covariance matrices are simply statistics of the forward process results, and these covariance matrices are shared among realizations, which simplifies the backward process.

The ENN is not sensitive to the size of the training dataset, which constitutes another advantage of this method over methods that are capable to perform uncertainty quantification, such as the Bayesian neural networks. This characteristic is especially important for real-world engineering problems where data collection is always expensive and time-consuming. It is even more desirable for the application of deep learning, in which tens of thousands of weights are used. This advantage is reflected in Fig. 8, in which the experiment shows that the ENN has higher prediction accuracy when the ratio of the amount of training data to the number of weights is small. This characteristic results from the ensemble of realizations generated by the RML method. The essential idea of the ENN is extracting more information from the training data. The training dataset is actually enlarged by the duplication with random disturbance among different realizations, which is similar to the commonly-used data augmentation methods in image recognition, such as rotation and adding noise. The disturbed realizations in the ensemble are helpful to avoid the over-fitting problem, as well. Although the ENN has the potential to solve problems with small datasets, it is not only for small data problems. The ENN is also able to provide predictions with large training datasets.

The ENN method is based on statistical analysis and the gradient is not required, which is different with the commonly-used networks based on backpropagation. This characteristic is not critical for traditional neural networks, however, because the derivatives of activation functions can be easily determined. The MCP model in traditional neural networks is coarse from a biological perspective, and it is replaced by the more elaborate and biomimetic HH model in neuroscience. However, the calculation of the HH model is complicated, which makes the frequently-used activation functions impracticable. The HH model requires more complicated activation functions, in which it is probably challenging to obtain the derivatives. The ENN is also a gradient-free method, which makes it preferable for neural networks based on the HH model in brain-inspired computing. This feature might be useful in the field of natural language processing, as well.

Three groups of computational experiments are carried out to test the performance of the ENN in this study. The first experiment shows the estimation and uncertainty quantification process in the ENN. In the second experiment, an ideal dataset is utilized to reveal the influences of the scale of measurement errors and the size of the training dataset. In the last experiment, the generalization capabilities of the ENN and three kinds of BNN methods are compared based on four real-world datasets. The results demonstrate that the ENN produces higher estimation accuracy than the BNN algorithms. The ENN is also robust to measurement error, and it works well for small data size problems. In conclusion, the ENN is a gradient-free method with the ability to perform uncertainty quantification, and it does well with small datasets. In addition to the batch learning mode shown in this study, the ENN can be casted as sequential (online) learning or mini-batch learning. Furthermore, the EnRML embedded in the ENN can be conveniently parallelized.

**Acknowledgements**

This work is partially funded by the National Natural Science Foundation of China (Grant No. U1663208 and 51520105005) and the National Science and Technology Major Project of China (Grant No. 2017ZX05009-005 and 2016ZX05037-003).

**Appendix A. Calculation of posterior probability distribution of the model parameters**

Assuming that the observations can be defined as the sum of estimation results and normally-distributed stochastic errors (Eq. (A.1)), the likelihood function is simply equivalent to the probability of the stochastic error, which is shown in Eq. (A.2):

$$d_{obs} = g(m) + \varepsilon \tag{A.1}$$

$$p(d_{obs}/m) = p(d_{obs} - g(m)) = p(\varepsilon) \tag{A.2}$$

where $g(m)$ is a function that maps $m$ to an estimation value; and $\varepsilon$ is a normally-distributed random vector with mean 0 and covariance matrix $C_D$.

Since the errors are normally distributed, $p(\varepsilon)$ can be determined by the probability distribution function of a multivariate normal distribution as:

$$p(\varepsilon) \propto \exp[-\frac{1}{2}(\varepsilon-0)^T C_D^{-1}(\varepsilon-0)] \tag{A.3}$$

Using Eq. (A.1) and Eq. (A.2), we can rewrite Eq. (A.3) and determine the likelihood function as:

$$p(d_{obs}/m) \propto \exp[-\frac{1}{2}(d_{obs} - g(m))^T C_D^{-1}(d_{obs} - g(m))] \tag{A.4}$$

The prior can be obtained similarly with the assumption that the model parameters are Gaussian variables:

$$p(m) \propto \exp[-\frac{1}{2}(m - m_{pr})^T C_M^{-1}(m - m_{pr})] \tag{A.5}$$

where $m_{pr}$ denotes the prior estimate of the model parameters; and $C_M$ denotes the prior covariance of the model parameters.

The posterior probability distribution of the model parameters can be obtained by the Bayesian theorem by multiplying the likelihood function (Eq. (A.4)) and the prior probability (Eq. (A.5)) as:

$$p(m/d_{obs}) \propto p(d_{obs}/m)p(m)$$
$$\propto \exp[-\frac{1}{2}(g(m)-d_{obs})^T C_D^{-1}(g(m)-d_{obs}) - \frac{1}{2}(m-m_{pr})^T C_M^{-1}(m-m_{pr})] \quad (A.6)$$
$$\propto \exp[-O(m)]$$

where $O(m)$ is the objective function, proportional to the posterior probability of the model parameters.

**Appendix B. Method for determining the multiplier $\lambda$**

The multiplier $\lambda$ is used to mitigate the influence of large data mismatch in early iterations. The determination method of $\lambda$ is based on the Levenberg-Marquardt algorithm (Chen & Oliver, 2013). The starting value of $\lambda$ should be on the same order of magnitude as, or lower than, $S_d(m_0)/(2N_d)$, where $m_0$ is the initial model parameters and $S_d(m_0)$ describes the data mismatch, given as:

$$S_d(m) = (g(m)-d_{obs})^T C_D^{-1}(g(m)-d_{obs}) \quad (B.1)$$

The multiplier $\lambda$ changes according to the performance of each iteration. If both the mean and the standard deviation of $S_d(m)$ of the ensemble are reduced, the value of $\lambda$ is reduced by a factor of $\gamma$. If the mean is reduced, but the standard deviation is increased, we accept the iteration and keep the value of $\lambda$ unchanged. If the mean of $S_d(m)$ of the ensemble is increased at one iteration, we reject the update and increase the value of $\lambda$ by a factor of $\gamma$, and repeat the current iteration with the larger value of $\lambda$. Essentially, this adjustment reduces the update step-size to find a better update, and it typically occurs at later iterations. It is necessary to set a lower bound of $\lambda$ because the update step-size cannot be too large, and it does not take many steps when the value of $\lambda$ needs to be increased. The lower bound of $\lambda$ is set to be 0.005 in this study, and the $\gamma$ is set to be 10.

**Appendix C. Two equivalent equations to change the dimension of the inverse matrix**

The derivation process of the two equivalent equations in Eq. (6) and Eq. (7) is shown as follows (Golub & Van Loan, 1989; Oliver et al., 2008):
Firstly, we can construct the following identical equations and rearrange them to obtain:

$$G^T + (G^T C_D^{-1})(GC_M G^T) = G^T C_D^{-1} C_D + (G^T C_D^{-1})(GC_M G^T)$$
$$= G^T C_D^{-1}(C_D + GC_M G^T) \quad (C.1)$$

$$G^{\mathrm{T}}+G^{\mathrm{T}}C_{\mathrm{D}}^{-1}G(C_{\mathrm{M}}G^{\mathrm{T}})=C_{\mathrm{M}}^{-1}(C_{\mathrm{M}}G^{\mathrm{T}})+G^{\mathrm{T}}C_{\mathrm{D}}^{-1}G(C_{\mathrm{M}}G^{\mathrm{T}})$$
$$=(C_{\mathrm{M}}^{-1}+G^{\mathrm{T}}C_{\mathrm{D}}^{-1}G)C_{\mathrm{M}}G^{\mathrm{T}} \tag{C.2}$$

The left-hand sides of Eq. (C.1) and Eq. (C.2) are identical. Thus, the right-hand sides must be equal as well, which gives:

$$G^{\mathrm{T}}C_{\mathrm{D}}^{-1}(C_{\mathrm{D}}+GC_{\mathrm{M}}G^{\mathrm{T}})=(C_{\mathrm{M}}^{-1}+G^{\mathrm{T}}C_{\mathrm{D}}^{-1}G)C_{\mathrm{M}}G^{\mathrm{T}} \tag{C.3}$$

The terms $(C_{\mathrm{D}}+GC_{\mathrm{M}}G^{\mathrm{T}})$ and $(C_{\mathrm{M}}^{-1}+G^{\mathrm{T}}C_{\mathrm{D}}^{-1}G)$ are both nonsingular positive-definite matrices. Thus, Eq. (C.3) can be rewritten as Eq. (C.4), which is same as Eq. (7):

$$(C_M^{-1}+G_l^T C_D^{-1} G_l)^{-1} G_l^T C_D^{-1} = C_M G_l^T (C_D + G_l C_M G_l^T)^{-1} \tag{C.4}$$

Regarding the derivation process of Eq. (6), we can construct the following identical equation:

$$C_{\mathrm{M}}=(C_{\mathrm{M}}^{-1}+G_l^{\mathrm{T}}C_{\mathrm{D}}^{-1}G_l)^{-1}(C_{\mathrm{M}}^{-1}+G_l^{T}C_{\mathrm{D}}^{-1}G_l)C_{\mathrm{M}}$$
$$=(C_{\mathrm{M}}^{-1}+G_l^{\mathrm{T}}C_{\mathrm{D}}^{-1}G_l)^{-1}C_{\mathrm{M}}^{-1}C_{\mathrm{M}}+(C_{\mathrm{M}}^{-1}+G_l^{\mathrm{T}}C_{\mathrm{D}}^{-1}G_l)^{-1}G_l^{T}C_{\mathrm{D}}^{-1}G_lC_{\mathrm{M}} \tag{C.5}$$

Using Eq. (C.4), we can rewrite Eq. (C.5) as:

$$C_{\mathrm{M}}=(C_{\mathrm{M}}^{-1}+G_l^{\mathrm{T}}C_{\mathrm{D}}^{-1}G_l)^{-1}+C_{\mathrm{M}}G_l^T(C_{\mathrm{D}}+G_lC_{\mathrm{M}}G_l^T)^{-1}G_lC_{\mathrm{M}} \tag{C.6}$$

Eq. (6) in section 2.2 can be obtained by simply rearranging terms in Eq. (C.6).

**Appendix D. Two approximations to substitute covariance for the gradients**

The derivation process of the two approximations in Eq. (10) and Eq. (11) is shown as follows (Reynolds et al., 2006; Li & Reynolds, 2007; Le et al., 2016):

Firstly, using first-order Taylor series to expand $m_j$ at $\bar{m}$ gives the following result:

$$g(m_j) = g(\bar{m}) + \bar{G}(m_j - \bar{m}) + e_j \tag{D.1}$$

where $\bar{G}$ is the sensitivity matrix for output as a function of $m$ evaluated at $\bar{m}$. Letting $\bar{e}$ denote the average of the residual error $e_j$ and taking averages in Eq. (D.1) gives:

$$\bar{g}(m) = g(\bar{m}) + \bar{e} \tag{D.2}$$

Subtracting Eq. (D.2) from Eq. (D.1) for $j = 1, ..., N_e$ gives:

$$g(m_j) - \bar{g}(m) = \bar{G}(m_j - \bar{m}) + (e_j - \bar{e}) \tag{D.3}$$

Secondly, the covariance between model parameters $m$ and estimation values $g(m)$, and the covariance of estimation values $g(m)$, respectively, are given by:

$$C_{M,D} = \frac{1}{N_e - 1} \sum_{j=1}^{N_e} (m_j - \bar{m})(g(m_j) - \bar{g}(m))^T \tag{D.4}$$

$$C_D = \frac{1}{N_e - 1} \sum_{j=1}^{N_e} (g(m_j) - \bar{g}(m))(g(m_j) - \bar{g}(m))^T \tag{D.5}$$

Thirdly, using Eq. (D.3), we can rewrite Eq. (D.4) and Eq. (D.5) as:

$$C_{M,D} = [\frac{1}{N_e - 1} \sum_{j=1}^{N_e} (m_j - \bar{m})(m_j - \bar{m})^T] \bar{G}^T + \frac{1}{N_e - 1} \sum_{j=1}^{N_e} (m_j - \bar{m})(e_j - \bar{e})^T \tag{D.6}$$

$$\begin{aligned} C_D = &\bar{G}(\frac{1}{N_e - 1} \sum_{j=1}^{N_e} (m_j - \bar{m})(m_j - \bar{m})^T) \bar{G}^T \\ &+ \frac{2}{N_e - 1} (\sum_{j=1}^{N_e} (m_j - \bar{m})(e_j - \bar{e})^T) \bar{G}^T + \frac{1}{N_e - 1} \sum_{j=1}^{N_e} (e_j - \bar{e})(e_j - \bar{e})^T \end{aligned} \tag{D.7}$$

Note that the first summation in Eq. (D.6) and Eq. (D.7) are the covariance matrix $C_M$. The second term in Eq. (D.6) and the last two terms in Eq. (D.7) are neglectable since the residual error $e_j$ from the first-order Taylor series is small. Thus, Eq. (D.6) and Eq. (D.7) can be reformulated as the following approximations (Zhang, 2001; Reynolds et al., 2006):

$$C_{M,D} \approx C_M \bar{G}^T \tag{D.8}$$

$$C_D \approx \bar{G} C_M \bar{G}^T \tag{D.9}$$

**Appendix E. Comparisons of the ENN and other methods**

*E.1 The difference between the BNN and the ENN*

Bayesian neural networks (BNN) is a kind of method that can obtain model uncertainty by giving distributions over the weights and biases in common neural networks. This method has been studied by Mackay (1992) and further extended by Neal (2012) and Gal (2016). The ENN proposed in this study seems similar to the BNN in terms of model structure. The objective function of the ENN is shown in Eq. (2), and the objective function of the BNN is given by:

$$O(m) = \frac{\beta}{2} \sum_{n=1}^{N} (g(m, x_n) - t_n)^2 + \frac{\alpha}{2} m^T m \tag{E.1}$$

where $N$ denotes the number of data points; $t_n$ denotes the target value of the n$^{th}$ data point; and $\alpha$ and $\beta$ are hyperparameters, which will be discussed later.

The objective functions of the ENN and the BNN are essentially the same equation based on the Bayesian theorem. However, the minimization method, determination of hyperparameters, and the method for solving the Hessian matrix are totally different.

*E.1.1 Minimization method of the ENN and the BNN*

Regarding the minimization method for the objective function, the ENN and the BNN are different. The ENN is based on the Gauss-Newton method using the calculation of the Hessian matrix. However, the BNN relies on the gradient descent method based on error backpropagation. Specifically, both the ENN and the BNN are iterative methods and obey:

$$m^{l+1} = m^l + \mu_l \delta m^{l+1} \tag{E.2}$$

where $l$ denotes the iteration step index; $\mu_l$ denotes the learning rate; and $\delta m^{l+1}$ denotes the modification value at this iteration step.

However, the ENN determines the modification value $\delta m^{l+1}$ by (Oliver et al., 2008):

$$\delta m^{l+1} = -H(m^l)^{-1}[C_M^{-1}(m^l - m_{pr}) + G_l^T C_D^{-1}(g(m^l) - d_{obs})] \tag{E.3}$$

and the BNN obtains this value based on the Delta rule, which is shown as (Rumelhart et al., 1986):

$$\delta m^{l+1} = \frac{\partial O(m)}{\partial m_{ij}^l} = \delta_j h(a_i)$$

$$\text{where} \begin{cases} \delta_j = h'(a_j) \sum_k (\delta_k m_{j,k}) & \text{for hidden layer} \\ \delta_j = -(d_{obs} - g(m^l)) & \text{for output layer} \end{cases} \quad \text{(E.4)}$$

where $H(m^l)$ is a Hessian matrix, which will be discussed later; $m_{ij}^l$ is the weight between two neurons from two adjacent layers; the input layer of $m_{ij}^l$ and the output layer of $m_{ij}^l$ denote two adjacent layers (which is different from the output layer of the whole neural network); $i$, $j$ and $k$, respectively, represent the i[th] element in the input layer of $m_{ij}^l$, the j[th] element in the output layer of $m_{ij}^l$, and the k[th] element in the layer after the output layer of $m_{ij}^l$; $a_i$ is the input to the activation function of the i[th] neuron; $h(a_i)$ means the activation of the j[th] neuron; and $h'(a_j)$ is the derivative of the activation function with respect to $a_j$. When the output layer of the weight $m_{ij}^l$ is the output of the neural networks and mean squared error (MSE) is used as a loss function, $\delta_j$ is the difference between the output value and the target value.

In conclusion, the BNN is based on gradient descent. This method possesses a hierarchical structure, which means that the gradient calculation of a weight requires all of the gradients of related weights in the deeper layers. This hierarchical structure results in the neurons in deeper neural networks requiring more complex calculations. In the ENN, however, there is no hierarchical structure in the calculation of the covariance matrix. Different weights have the same computational complexity to determine its covariance.

*E.1.2 Determination of hyperparameters of the ENN and the BNN*

Hyperparameters are utilized to balance the influence of model mismatch and data mismatch. The hyperparameters in the ENN are $C_D$ and $C_M$, which are, respectively, equivalent to the covariance matrix of the stochastic error and the covariance of the prior weights. The hyperparameters in the ENN are fixed and known from the beginning; whereas, the BNN requires an iterative procedure to determine the hyperparameters, which is shown as (Bishop, 2007):

$$\alpha = \frac{\gamma}{m_{MAP}^T m_{MAP}} \quad \text{(E.5)}$$

$$\beta = [\frac{1}{N-\gamma}\sum_{n=1}^{N}(y(x_n, m_{MAP})-t_n)^2]^{-1} \qquad (E.6)$$

$$\gamma = \sum_{i=1}^{W}\frac{\lambda_i}{\alpha+\lambda_i} \qquad (E.7)$$

$$\beta H u_i = \lambda_i u_i \qquad (E.8)$$

where $m_{MAP}$ is the neural networks' weights that maximize the posterior $p(m|d_{obs})$, which is equivalent to minimizing the objective function Eq. (E.1); $\gamma$ denotes the effective number of parameters and is defined by Eq. (E.7); $W$ is the total number of parameters in the neural networks' weights; and $\lambda_i$ are eigenvalues determined by the eigenvalue equation in Eq. (E.8).

Note that the solution for $\alpha$ and $\beta$ in Eq. (E.5) and Eq. (E.6), respectively, is implicit because the effective number of parameters $\gamma$ and weights $m_{MAP}$ depend on $\alpha$ and $\beta$ themselves. Therefore, an iterative procedure is adopted in which we make an initial choice for $\alpha$ and $\beta$, and use them to find $m_{MAP}$, which is given by Eq. (E.2), and also calculate $\gamma$ according to Eq. (E.7) and Eq. (E.8). These values are then used to re-estimate $\alpha$ and $\beta$. This process is repeated until convergence (Bishop, 2007).

*E.1.3 Method for solving the Hessian matrix of the ENN and the BNN*

Calculation of the Hessian matrix constitutes another problem for both methods. The ENN needs the Hessian matrix to calculate $\delta m^{l+1}$ in Eq. (E.3), and the BNN utilizes it to determine hyperparameters based on Eq. (E.8). The solving methods are different between the ENN and the BNN not only in the calculation processes, but also in the definition of the Hessian matrix. In the ENN, the Hessian matrix comprises the second derivatives of the objective function Eq. (3) with respect to the components of $m$; whereas, the derivative object of the Hessian matrix is not the objective function, but the sum-of-squares error function in the BNN (Bishop, 2007). Regarding the calculation of the Hessian matrix in the ENN, the second derivatives of $g(m)$ are ignored according to the Gauss-Newton method, which is shown as (Oliver et al., 2008):

$$\begin{aligned}H &= \nabla((\nabla O(m))^T) = \nabla[C_m^{-1}(m-m_{pr})+G^T C_D^{-1}(d_{obs}-g(m))] \\ &= C_M^{-1}+G^T C_D^{-1}G+(\nabla G^T)C_D^{-1}(d_{obs}-g(m)) \\ &\approx C_M^{-1}+G^T C_D^{-1}G\end{aligned} \qquad (E.9)$$

In the BNN, diagonal approximation and outer product approximation are the most commonly-used methods to calculate the Hessian matrix (Eade & Robb, 1981; Bishop, 2007). The diagonal approximation is shown in Eq. (E.10). It is based on the error backpropagation method with the assumption that the off-diagonal elements in the second-derivative terms are neglectable. However, in practice, the Hessian is typically found to be strongly nondiagonal, which is the major problem with this method.

$$\mathbf{H} = \frac{\partial^2 E}{\partial m_{i,j}^2} = \frac{\partial^2 E}{\partial a_j^2} h(a_i)^2 \approx [h'(a_j)^2 \sum_k m_{k,j}^2 \frac{\partial^2 E_n}{\partial a_k^2} + h''(a_j) \sum_k m_{k,j} \frac{\partial E}{\partial a_k}] h(a_i)^2 \qquad (E.10)$$

The outer product approximation is shown in Eq. (E.11). This approximation neglects the second terms, as well. The limitation of this method is that it is only likely to be valid for weights near the minimum, or when the network has already been trained appropriately.

$$\mathbf{H} = \nabla\nabla E = \sum_{n=1}^{N} \nabla y_n \nabla y_n^{\mathrm{T}} + \sum_{n=1}^{N} (y_n - t_n) \nabla\nabla y_n \approx \sum_{n=1}^{N} \nabla y_n \nabla y_n^{\mathrm{T}} \qquad (E.11)$$

*E.2 Difference between the Monte Carlo and the ENN*

Monte Carlo relies on repeated random sampling to perform uncertainty quantification. For example, the Monte Carlo estimate is used in the BNN, and in practice the Monte Carlo estimate is equivalent to performing several stochastic forward passes through the network and averaging the results (Gal, 2016). Although the ENN method also has stochastic realizations that pass through the network, it is essentially different from the Monte Carlo estimate.

In the Monte Carlo estimate, each realization has an independent forward process and an independent backward process for each iteration. In contrast, realizations in the ENN share information. As shown in Eq. (10), the covariance matrix $C_{M_l}$ and $C_{D_l}$, and $C_{M_l,D_l}$ are used to calculate the updated weights. The covariance matrices are all statistical variables obtained from the ensemble of realizations, which means that each realization is an indispensable part of the ensemble to estimate the weights in the ENN. In other words, the forward process in the ENN can not only obtain estimation values, but also obtain the shared covariance matrices for update, which simplifies calculation in the backward process.

*E.3 Difference between ensemble averaging and the ENN*

Ensemble averaging is a learning paradigm in which alternative proposals, called components, combine their individual outputs to produce a unique solution to a given problem. In practice, the components are multiple independent models, and average their predictions at test time. The neural networks always exhibit better improvement with higher model variety in the ensemble (Jiménez, 1998; Zhou et al., 2002). However, the word "ensemble" in the ENN does not hold the same meaning as the aforementioned ensemble method. The ensemble in the ENN comes from the original meaning of this word. In this study, the ensemble represents a set of realizations of the same distribution that go together to form a whole. In other words, the components in the ensemble are not independent models, but different realizations of dataset and model parameters. The dataset in each realization is sampled from the training dataset by the RML method, in which the mean of the probability distribution equals the observed target data in the training dataset, and the variance is determined by the estimation of measurement error. The model parameters are sampled from the prior probability distribution in the same method.

**Appendix F. The PRES-2D dataset**

In this problem, a 2D porous domain with a size of 1200×1200 ft is examined, and the domain is simulated by a 40×40 grid. The porous media is saturated with oil initially, and water is injected at the point with coordinates (1, 1) of the grid with a constant injection rate. The shock wave of the water flooding front will propagate through the domain, and oil is produced at (40, 40). The two fluids are immiscible, and the saturations of irreducible water and residual oil are $S_{wi} = 0.2$ and $S_{or} = 0.2$, respectively. Capillary pressure and gravity are neglected for simplicity. The permeability field is randomly generated by the K-L expansion (Li & Zhang, 2007), where 20 eigenfunctions are multiplied by 20 random variables. The eigenfunctions are illustrated in Fig. F.1, and several examples of permeability fields generated with different sets of 20 random variables are shown in Fig. F.2. In the PRES-2D, the input of each data point consists of a set of the 20 random variables, which generate a corresponding permeability field by the K-L expansion. The permeability field determines the pressure field of the domain according to complex physic laws. The outputs of the PRES-2D are the pressure values at five different observation points. The determination of the pressure field is a highly nonlinear problem, and it is difficult to obtain an analytical solution. Thus, a commonly-used underground flow simulator, called Eclipse, is employed to simulate the pressure field. Finally, the dependent variables in PRES-2D are the values of flow pressure, and the independent variables are sets of 20 random variables that are utilized to generate the random permeability fields. The PRES-2D dataset is provided in .txt format in the Supplementary Materials.

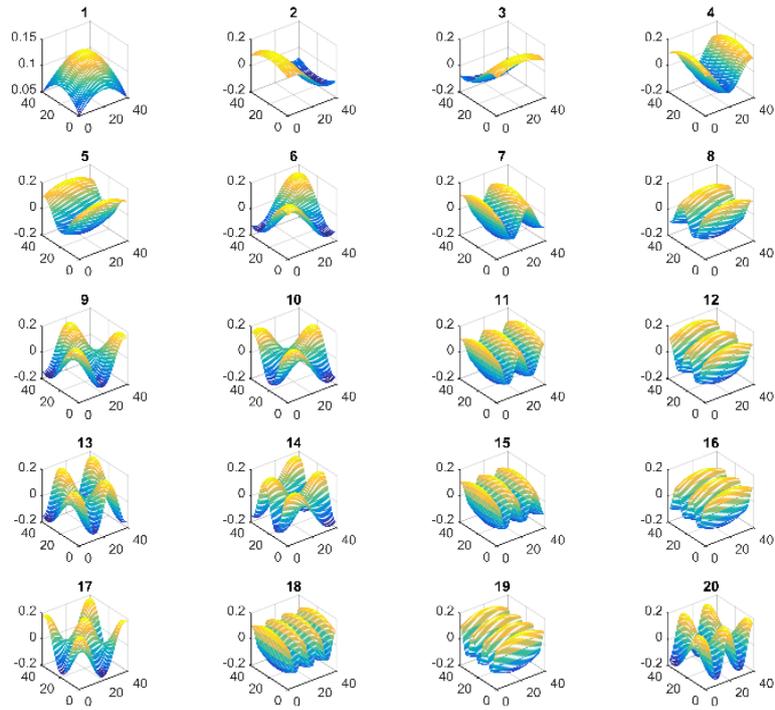

**Fig. F.1.** Illustration of the 20 eigenfunctions in the K-L expansion. The stochastic permeability fields are generated based on the products of the eigenfunctions and the random variables.

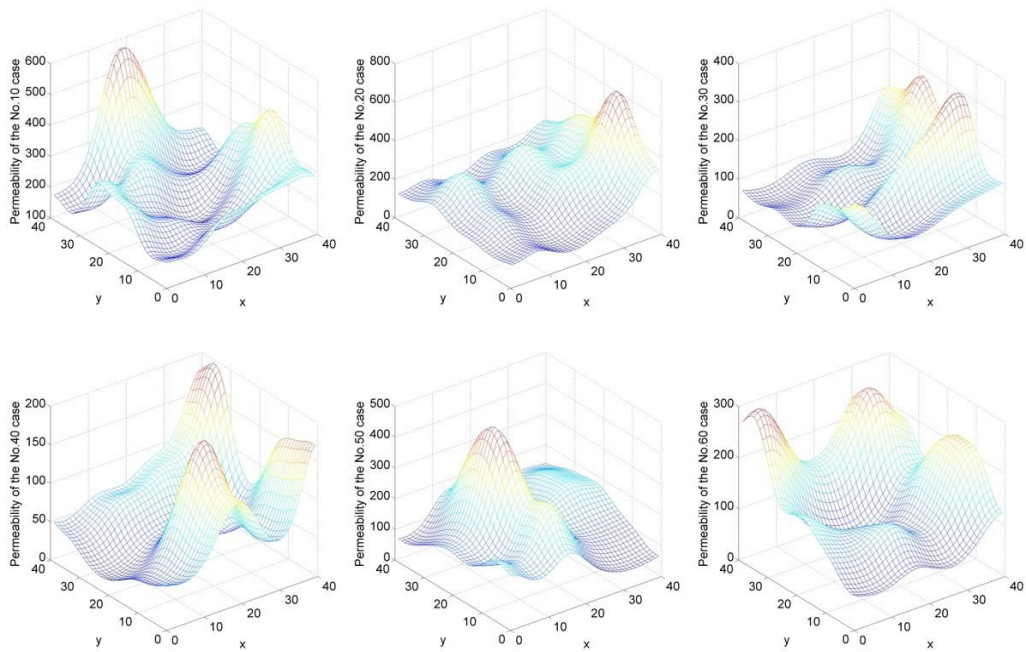

**Fig. F.2.** Examples of the random permeability fields. Each field is generated based on the eigenfunctions in Fig. F1 with a different set of random variables according to the K-L expansion.

In *Neural Networks Proceedings, 1998. IEEE World Congress on Computational Intelligence. The 1998 IEEE International Joint Conference on* (Vol. 1, pp. 753-756). IEEE.

Kaya, H., Tüfekci, P., & Gürgen, F. S. (2012, March). Local and global learning methods for predicting power of a combined gas & steam turbine. In *Proceedings of the International Conference on Emerging Trends in Computer and Electronics Engineering* (pp. 13-18).

Kim, K. K. K., Patrón, E. R., & Braatz, R. D. (2018). Standard representation and unified stability analysis for dynamic artificial neural network models. *Neural Networks*, *98*, 251-262.

Lakshminarayanan, B., Pritzel, A., & Blundell, C. (2017). Simple and scalable predictive uncertainty estimation using deep ensembles. In *Advances in Neural Information Processing Systems* (pp. 6405-6416).

Le, D. H., Emerick, A. A., & Reynolds, A. C. (2016). An adaptive ensemble smoother with multiple data assimilation for assisted history matching. *SPE Journal*, *21*(06), 2-195.

Li, F. F., Karpathy, A., & Johnson, J. (2015). CS231n: Convolutional neural networks for visual recognition. *University Lecture*.

Li, G., & Reynolds, A. C. (2007, January). An iterative ensemble Kalman filter for data assimilation. In *SPE Annual Technical Conference and Exhibition*. Society of Petroleum Engineers.

Li, H., & Zhang, D. (2007). Probabilistic collocation method for flow in porous media: Comparisons with other stochastic methods. *Water Resources Research, 43*(9), 1-13.

Li, R., Reynolds, A.C., & Oliver, D.S. (2003). History matching of three-phase flow production data. *SPE Journal. 8* (4), 328–340.

Lichman, M. (2013). UCI Machine Learning Repository [http://archive.ics.uci.edu/ml]. Irvine, CA: University of California, School of Information and Computer Science.

Louizos, C., & Welling, M. (2017). Multiplicative normalizing flows for variational Bayesian neural networks. *arXiv preprint arXiv:1703.01961*.

MacKay, D. J. (1992). A practical Bayesian framework for backpropagation networks. *Neural Computation*, *4*(3), 448-472.

McCulloch, W. S., & Pitts, W. (1943). A logical calculus of the ideas immanent in nervous activity. *The Bulletin of Mathematical Biophysics*, *5*(4), 115-133.

Naftaly, U., Intrator, N., & Horn, D. (1997). Optimal ensemble averaging of neural networks. *Network: Computation in Neural Systems, 8*(3):283–296.

Neal, R. M. (2012). *Bayesian learning for neural networks* (Vol. 118). Springer Science & Business Media.

Oliver, D. S., Reynolds, A. C., & Liu, N. (2008). *Inverse theory for petroleum reservoir characterization and history matching*. Cambridge University Press.

Papineni, K., Roukos, S., Ward, T., & Zhu, W. J. (2002, July). BLEU: a method for automatic evaluation of machine translation. In *Proceedings of the 40th Annual Meeting on Association for Computational Linguistics* (pp. 311-318). Association for Computational Linguistics.

Quinlan, J. R. (1993). Combining instance-based and model-based learning. In *Proceedings of the Tenth International Conference on Machine Learning* (pp. 236-243).

Reiter, E. (2018). A structured review of the validity of BLEU. *Computational Linguistics*, *44*(3), 1-12.

Reynolds, A. C., Zafari, M., & Li, G. (2006, September). Iterative forms of the ensemble Kalman filter. In *ECMOR X-10th European Conference on the Mathematics of Oil Recovery*.

Rumelhart, D. E., Hinton, G. E., & Williams, R. J. (1986). Learning representations by back-

**Supplementary Material.** Detailed example of the calculation process of the ENN

To demonstrate the ENN more clearly, an end-to-end example worked out by hand is provided in this Supplementary Material. A two-layer neural network with five neurons in the hidden layer is trained to simulate the mapping of y = 2x+1. All of the weights, the covariance matrices, and the estimations in the first three iterations are provided in a step by step method to show the details of the ENN algorithm. The estimation value, the estimation loss, the $\lambda$ (step size), and the weights at the first 31 steps (the initial step and 30 iteration steps) are shown in figures to demonstrate the converging process.

In this demonstration, the ensemble size is 10, which means that the covariance matrices are calculated based on 10 different realizations. The training dataset includes six samples. The inputs of these training samples are 1, 2, 4, 6, 8, and 9, and the corresponding outputs are 3, 5, 9, 13, 17, and 19, respectively. The inputs of the test data are 3, 5, and 7, and the corresponding target values are 7, 11, and 15, respectively. There are 16 different weights in this neural network and only six training samples. Although the ratio of the number of training data to the number of weights is only 0.375, the ENN is still able to simulate the mapping.

*Step 1 (initial state)*

First, a 16×10 matrix $m^1$ is randomly generated based on a standard normal distribution $\mathcal{N}(0,1)$ as the initial weights of the 10 realizations. Each column of the $m^1$ represents the 16 weights of the same realization of the neural network. Each row of the $m^1$ represents the 10 realizations of a certain weight. The randomly generated $m^1$ is shown as follows:

$m^1 =$

$$\begin{bmatrix}
0.383 & 0.448 & 0.406 & -0.337 & 0.634 & 1.023 & -2.928 & -1.307 & -1.893 & 1.614 \\
1.119 & -0.408 & 0.549 & 0.380 & 0.522 & -1.571 & 2.586 & -1.229 & 1.276 & -0.578 \\
-0.423 & 0.476 & -0.871 & -1.024 & -0.291 & -0.315 & -1.247 & -0.590 & -0.346 & -0.237 \\
-0.536 & -1.904 & 0.906 & 0.465 & -0.282 & 0.940 & -0.019 & -0.160 & -0.614 & -0.199 \\
-0.315 & 1.008 & 0.184 & -0.272 & -0.041 & 0.873 & 0.229 & 0.039 & -1.415 & 0.471 \\
1.471 & -0.495 & -0.108 & -0.182 & -0.499 & 1.364 & -0.550 & 0.813 & 1.786 & 1.472 \\
1.697 & 1.177 & -0.632 & -0.593 & -1.340 & -1.151 & 0.158 & 0.529 & -0.569 & -0.865 \\
0.933 & -0.890 & 0.359 & 0.065 & -0.993 & -1.283 & 0.420 & 0.158 & 0.762 & -0.636 \\
-2.486 & 0.821 & -0.043 & 1.341 & -0.095 & 0.060 & 0.495 & -0.384 & -0.144 & -0.079 \\
-1.846 & 0.154 & 0.425 & 0.705 & -3.570 & 0.177 & -0.362 & -0.725 & 0.685 & -0.140 \\
1.057 & -1.339 & -0.124 & -0.954 & -0.127 & 0.193 & -0.606 & 0.213 & -1.258 & 0.567 \\
-0.555 & -1.602 & -1.751 & -0.909 & 0.482 & -0.116 & 1.039 & 0.337 & 1.093 & -1.102 \\
1.051 & 0.527 & -1.029 & -1.713 & -0.072 & 1.187 & 0.715 & 1.005 & -1.789 & -0.705 \\
-1.096 & -0.362 & 0.679 & -0.004 & -1.006 & -0.814 & 2.016 & -0.078 & -0.464 & 1.105 \\
-1.584 & 0.828 & 0.158 & 0.005 & -1.699 & 0.567 & -0.752 & -0.474 & 0.587 & -0.182 \\
0.838 & -0.987 & -1.156 & -0.485 & 1.301 & 0.524 & -0.182 & -1.809 & -1.459 & -0.395
\end{bmatrix}$$

The estimations of the training data $g(m^1)_{\text{train}}$ of the 10 realizations are obtained based on the initial weight matrix $m^1$ and the inputs of the training data $X_{\text{train}} = [1, 2, 4, 6, 8, 9]$, which is shown as follows:

$g(m^1)_{\text{train}} = f(m^1, X_{\text{train}}) =$

$$\begin{bmatrix} 3.569 & 1.158 & -1.509 & 0.671 & 3.666 & 2.457 & -0.139 & -1.204 & -2.016 & -1.142 \\ 3.547 & 0.532 & -0.764 & 1.696 & 3.583 & 2.733 & 0.175 & -1.238 & 0.622 & -0.915 \\ 3.134 & -1.179 & -0.740 & 2.609 & 3.488 & 2.974 & 0.517 & -1.216 & 1.018 & -0.659 \\ 2.517 & -2.493 & -0.961 & 2.944 & 3.457 & 3.081 & 0.895 & -1.181 & 0.861 & -0.520 \\ 2.116 & -2.861 & -1.096 & 3.052 & 3.450 & 3.150 & 1.128 & -1.151 & 0.706 & -0.451 \\ 2.016 & -2.906 & -1.140 & 3.074 & 3.449 & 3.176 & 1.198 & -1.137 & 0.641 & -0.431 \end{bmatrix}$$

where $f$ denotes the mapping function defined by the neural network. Each column of the $g(m^1)_{\text{train}}$ represents the six outputs of the training data in the same realization, and the 10 columns show the results of the 10 realizations.

The final estimation values of the training dataset $Y_{\text{train},1}$ is obtained by averaging the $g(m^1)_{\text{train}}$ along its rows. It is obvious that the estimation $Y_{\text{train},1}$ is different with the target values $T_{\text{train}}$ since the weights are randomly generated.

$Y_{\text{train},1} =$
$$\begin{bmatrix} 0.551 & 0.997 & 0.995 & 0.860 & 0.804 & 0.794 \end{bmatrix}^{\text{T}}$$

$T_{\text{train}} =$
$$\begin{bmatrix} 3.000 & 5.000 & 9.000 & 13.000 & 17.000 & 19.000 \end{bmatrix}^{\text{T}}$$

The estimation values of the test dataset $g(m^1)_{\text{test}}$ of the 10 realizations, the averaged estimation results $Y_{\text{test},1}$, and the target values of the test dataset $T_{\text{test}}$ are also obtained in the same method based on the inputs of the test dataset $X_{\text{test}} = [3, 5, 7]$.

$g(m^1)_{\text{test}} = f(m^1, X_{\text{test}}) =$
$$\begin{bmatrix} 3.394 & -0.278 & -0.635 & 2.260 & 3.523 & 2.893 & 0.312 & -1.233 & 1.034 & -0.771 \\ 2.819 & -1.978 & -0.861 & 2.820 & 3.468 & 3.034 & 0.721 & -1.198 & 0.945 & -0.577 \\ 2.279 & -2.750 & -1.038 & 3.013 & 3.452 & 3.119 & 1.029 & -1.165 & 0.780 & -0.480 \end{bmatrix}$$

$Y_{\text{test},1} =$
$$\begin{bmatrix} 1.050 & 0.919 & 0.824 \end{bmatrix}^{\text{T}}$$

$T_{\text{test}} =$
$$\begin{bmatrix} 7.000 & 11.000 & 15.000 \end{bmatrix}^{\text{T}}$$

Based on the averaged estimation results and the target values, the estimation loss is defined as Eq. (S.1):

$$L = \frac{\sum |Y - T|}{N} \tag{S.1}$$

where $Y$ and $T$ represent the averaged estimation results and the target values, respectively; and N equals to the number of data in $Y$ or $T$.

According to Eq. (S.1), the loss of the training dataset $L_{\text{train},1}$ and the loss of the test dataset $L_{\text{test},1}$ at step 1 are calculated and shown as follows:

$L_{\text{train},1} = 10.167$
$L_{\text{test},1} = 10.069$

The data mismatch of the model $S_d(m^1)$ is calculated based on the method introduced in Appendix B, and its mean $E(S_d(m^1))$ and its standard deviation $\sigma(S_d(m^1))$ are shown as follows:

$E(S_d(m^1)) = 212611550$

$\sigma(S_d(m^1)) = 61806331$

*Step 2*

At the following iteration step, a multiplier $\lambda$ is applied in the Hessian term in Eq. (10) to mitigate the influence of large data mismatch. $\lambda$ is inversely proportional to the step size in the iteration, which means that it will increase when large data mismatch is obtained and results in small step size. The method for determining $\lambda$ is introduced in Appendix B, and its value at step 2 is shown in the following equation. It should be mentioned that, although this $\lambda$ is used at step 2, 1 is taken as the subscript of $\lambda$ since it is calculated based on the results at step 1.

$\lambda_1 = 17717629$

The perturbed observation $D^1{}_{obs}$ is a matrix that includes all of the perturbed observations of different realizations. The j$^{th}$ column of the $D^1{}_{obs}$ represents the perturbed observation of the j$^{th}$ realization ($d_{obs,j}$). Each column of the $D^1{}_{obs}$ is independently sampled from a multivariate Gaussian distribution with mean $d_{obs}$ and covariance $C_D$. The perturbed observation at step 2 is given as:

$D^1{}_{obs} =$

$$\begin{bmatrix} 2.990 & 3.006 & 3.004 & 2.998 & 3.006 & 3.000 & 2.996 & 2.994 & 3.016 & 3.004 \\ 4.967 & 4.987 & 4.999 & 5.009 & 5.007 & 5.010 & 5.005 & 4.981 & 4.985 & 5.001 \\ 9.020 & 9.004 & 9.003 & 8.998 & 9.003 & 8.962 & 9.036 & 9.000 & 8.987 & 8.983 \\ 13.045 & 12.994 & 13.030 & 13.018 & 13.004 & 13.023 & 12.994 & 12.970 & 12.962 & 12.999 \\ 16.992 & 17.049 & 16.978 & 17.029 & 17.058 & 17.016 & 17.066 & 16.992 & 17.034 & 17.037 \\ 19.044 & 19.019 & 19.040 & 18.985 & 18.969 & 18.976 & 19.048 & 18.956 & 18.934 & 18.965 \end{bmatrix}$$

The $C_D$ denotes the covariance matrix of the normally-distributed stochastic errors in Eq. (2), which is a diagonal matrix with the diagonal elements of $0.002^2$. The $C_M$ denotes the covariance matrix of prior model parameters, which equals to an identity matrix.

The $C_{M_l,D_l}$ denotes the cross-covariance matrix between the updated model parameters and the predictions at iteration step *l* based on the ensemble of realizations. The $C_{D_l}$ and $C_{M_l}$ are the covariance matrix of the predicted data and the covariance matrix of the updated model parameters at iteration step *l*, respectively. The $C_{M_1,D_1}$, $C_{D_1}$, and $C_{M_1}$ are the matrices obtained based on the results of step 1.

$$C_{M_1,D_1} = \frac{(m^1 - \overline{m^1})(g(m^1)_{train} - \overline{g(m^1)_{train}})^T}{N_e - 1} =$$

$$\begin{bmatrix} 1.213 & 0.687 & 0.420 & 0.216 & 0.131 & 0.115 \\ -0.139 & 0.246 & 0.400 & 0.475 & 0.479 & 0.473 \\ 0.262 & 0.127 & -0.128 & -0.313 & -0.375 & -0.386 \\ -0.162 & 0.015 & 0.445 & 0.752 & 0.853 & 0.870 \\ 0.377 & -0.114 & -0.334 & -0.410 & -0.398 & -0.385 \\ -0.305 & 0.030 & 0.183 & 0.216 & 0.181 & 0.164 \\ 0.287 & -0.048 & -0.471 & -0.793 & -0.923 & -0.947 \\ -0.605 & -0.262 & -0.120 & -0.094 & -0.126 & -0.142 \\ -0.678 & -0.577 & -0.471 & -0.356 & -0.251 & -0.221 \\ -1.895 & -1.337 & -1.157 & -1.122 & -1.092 & -1.086 \\ 0.534 & 0.254 & 0.307 & 0.371 & 0.366 & 0.363 \\ -0.013 & 0.324 & 0.605 & 0.846 & 0.932 & 0.947 \\ 1.219 & 0.457 & 0.095 & -0.036 & -0.059 & -0.051 \\ -1.282 & -1.251 & -1.054 & -0.840 & -0.724 & -0.694 \\ -0.962 & -0.700 & -0.761 & -0.849 & -0.857 & -0.857 \\ 1.891 & 1.563 & 1.572 & 1.619 & 1.624 & 1.626 \end{bmatrix}$$

$$C_{D_1} = \frac{(g(m^1)_{train} - \overline{g(m^1)}_{train})(g(m^1)_{train} - \overline{g(m^1)}_{train})^T}{N_e - 1} =$$

$$\begin{bmatrix} 4.453 & 3.464 & 3.044 & 2.798 & 2.691 & 2.677 \\ 3.464 & 3.275 & 3.224 & 3.132 & 3.047 & 3.025 \\ 3.044 & 3.224 & 3.655 & 3.889 & 3.909 & 3.903 \\ 2.798 & 3.132 & 3.889 & 4.378 & 4.492 & 4.504 \\ 2.691 & 3.047 & 3.909 & 4.492 & 4.648 & 4.671 \\ 2.677 & 3.025 & 3.903 & 4.504 & 4.671 & 4.696 \end{bmatrix}$$

$$C_{M_1} = \frac{(m^1 - \overline{m^1})(m^1 - \overline{m^1})^T}{N_e - 1} =$$

$$\begin{bmatrix} 2.018 & -1.081 & 0.364 & 0.085 & 0.470 & 0.219 & -0.299 & -0.665 & -0.217 & -0.353 & 0.502 & -1.009 & 0.072 & -0.546 & 0.028 & 0.579 \\ -1.081 & 1.562 & -0.301 & -0.102 & -0.433 & -0.335 & 0.212 & 0.628 & -0.114 & -0.185 & -0.245 & 0.475 & -0.356 & 0.459 & -0.404 & 0.167 \\ 0.364 & -0.301 & 0.234 & -0.274 & 0.101 & 0.094 & 0.097 & -0.197 & -0.051 & -0.085 & -0.028 & -0.143 & 0.112 & -0.258 & 0.117 & 0.019 \\ 0.085 & -0.102 & -0.274 & 0.684 & -0.001 & 0.066 & -0.499 & 0.011 & 0.058 & 0.187 & 0.216 & 0.015 & -0.134 & 0.161 & -0.018 & 0.113 \\ 0.470 & -0.433 & 0.101 & -0.001 & 0.467 & -0.226 & 0.036 & -0.379 & 0.178 & 0.013 & 0.092 & -0.357 & 0.442 & 0.119 & 0.113 & 0.124 \\ 0.219 & -0.335 & 0.094 & 0.066 & -0.226 & 0.924 & 0.006 & 0.155 & -0.541 & 0.218 & 0.337 & 0.170 & -0.003 & -0.275 & 0.107 & -0.070 \\ -0.299 & 0.212 & 0.097 & -0.499 & 0.036 & 0.006 & 1.029 & 0.352 & -0.447 & -0.054 & 0.075 & -0.168 & 0.549 & -0.094 & -0.127 & -0.165 \\ -0.665 & 0.628 & -0.197 & 0.011 & -0.379 & 0.155 & 0.352 & 0.610 & -0.326 & 0.167 & 0.016 & 0.175 & -0.219 & 0.151 & -0.145 & -0.266 \\ -0.217 & -0.114 & -0.051 & 0.058 & 0.178 & -0.541 & -0.447 & -0.326 & 1.007 & 0.642 & -0.594 & -0.109 & -0.457 & 0.383 & 0.491 & -0.326 \\ -0.353 & -0.185 & -0.085 & 0.187 & 0.013 & 0.218 & -0.054 & 0.167 & 0.642 & 1.779 & -0.482 & -0.384 & -0.614 & 0.505 & 1.021 & -0.929 \\ 0.502 & -0.245 & -0.028 & 0.216 & 0.092 & 0.337 & 0.075 & 0.016 & -0.594 & -0.482 & 0.627 & -0.081 & 0.427 & -0.091 & -0.397 & 0.353 \\ -1.009 & 0.475 & -0.143 & 0.015 & -0.357 & 0.170 & -0.168 & 0.175 & -0.109 & -0.384 & -0.081 & 1.074 & 0.157 & -0.006 & -0.278 & 0.113 \\ 0.072 & -0.356 & 0.112 & -0.134 & 0.442 & -0.003 & 0.549 & -0.219 & -0.457 & -0.614 & 0.427 & 0.157 & 1.327 & -0.168 & -0.305 & 0.393 \\ -0.546 & 0.459 & -0.258 & 0.161 & 0.119 & -0.275 & -0.094 & 0.151 & 0.383 & 0.505 & -0.091 & -0.006 & -0.168 & 0.997 & 0.092 & -0.341 \\ 0.028 & -0.404 & 0.117 & -0.018 & 0.113 & 0.107 & -0.127 & -0.145 & 0.491 & 1.021 & -0.397 & -0.278 & -0.305 & 0.092 & 0.772 & -0.556 \\ 0.579 & 0.167 & 0.019 & 0.113 & 0.124 & -0.070 & -0.165 & -0.266 & -0.326 & -0.929 & 0.353 & 0.113 & 0.393 & -0.341 & -0.556 & 1.039 \end{bmatrix}$$

The updated model parameters ($m^2$) are calculated according to Eq. (10), based on the model parameters from the previous step ($m^1$), the estimation results from the previous step ($g(m^1)_{train}$), the step size ($\lambda_1$), the perturbed observation ($D^1_{obs}$), the cross-covariance matrix ($C_{M_1,D_1}$), and the covariance matrices ($C_{D_1}$, and $C_{M_1}$). The $m_{pr}$ denotes the prior estimate of the model parameters, which equals to the initial model parameters ($m^1$) if the prior information is not available. The $d_{obs,j}$ in Eq. (10) equals to the j$^{th}$ column in the perturbed observation matrix ($D^1_{obs}$). The ultimate result of model parameters at step 2 is shown as follows:

$m^2 =$

$$\begin{bmatrix}
0.442 & 0.601 & 0.605 & -0.217 & 0.688 & 1.105 & -2.768 & -1.107 & -1.710 & 1.807 \\
1.380 & -0.046 & 0.878 & 0.627 & 0.761 & -1.325 & 2.876 & -0.896 & 1.562 & -0.258 \\
-0.619 & 0.226 & -1.086 & -1.195 & -0.470 & -0.493 & -1.439 & -0.806 & -0.535 & -0.444 \\
-0.091 & -1.307 & 1.441 & 0.875 & 0.124 & 1.352 & 0.455 & 0.379 & -0.141 & 0.319 \\
-0.540 & 0.707 & -0.077 & -0.473 & -0.247 & 0.666 & -0.005 & -0.227 & -1.639 & 0.216 \\
1.580 & -0.353 & 0.010 & -0.090 & -0.399 & 1.462 & -0.442 & 0.935 & 1.886 & 1.588 \\
1.216 & 0.532 & -1.206 & -1.035 & -1.781 & -1.597 & -0.351 & -0.050 & -1.075 & -1.421 \\
0.882 & -0.989 & 0.241 & -0.012 & -1.040 & -1.342 & 0.324 & 0.041 & 0.652 & -0.749 \\
-2.622 & 0.584 & -0.296 & 1.171 & -0.220 & -0.086 & 0.281 & -0.640 & -0.368 & -0.326 \\
-2.398 & -0.718 & -0.465 & 0.080 & -4.074 & -0.384 & -1.121 & -1.623 & -0.113 & -1.002 \\
1.241 & -1.061 & 0.154 & -0.756 & 0.039 & 0.374 & -0.369 & 0.491 & -1.009 & 0.834 \\
-0.066 & -0.927 & -1.132 & -0.443 & 0.929 & 0.344 & 1.583 & 0.963 & 1.638 & -0.502 \\
0.996 & 0.511 & -0.988 & -1.707 & -0.123 & 1.159 & 0.736 & 1.043 & -1.742 & -0.667 \\
-1.481 & -0.989 & 0.035 & -0.450 & -1.359 & -1.210 & 1.467 & -0.731 & -1.034 & 0.478 \\
-2.022 & 0.171 & -0.489 & -0.461 & -2.099 & 0.136 & -1.309 & -1.126 & 0.009 & -0.808 \\
1.674 & 0.280 & 0.098 & 0.413 & 2.065 & 1.353 & 0.897 & -0.543 & -0.342 & 0.820
\end{bmatrix}$$

Based on the updated model parameters $m^2$ and the inputs of the training dataset $X_{\text{train}} = [1, 2, 4, 6, 8, 9]$, the estimations of the training dataset with 10 realizations are obtained as $g(m^2)_{\text{train}}$. The averaged estimation values of the training dataset $Y_{\text{train},2}$ are obtained by averaging the $g(m^2)_{\text{train}}$ along its rows.

$g(m^2)_{\text{train}} = f(m^2, X_{\text{train}}) =$

$$\begin{bmatrix}
4.251 & 1.055 & 0.205 & 1.087 & 5.308 & 3.782 & 0.209 & 0.338 & -1.171 & 1.249 \\
3.951 & 0.453 & 1.444 & 1.978 & 5.023 & 4.031 & -0.476 & -0.217 & 1.716 & 1.540 \\
2.996 & -0.550 & 2.083 & 2.667 & 4.795 & 3.821 & -1.806 & -0.723 & 1.828 & 1.762 \\
2.594 & -1.107 & 2.246 & 2.987 & 4.759 & 3.682 & -2.701 & -1.106 & 1.800 & 1.810 \\
2.533 & -1.512 & 2.370 & 3.159 & 4.755 & 3.649 & -3.156 & -1.357 & 1.796 & 1.818 \\
2.528 & -1.660 & 2.424 & 3.208 & 4.755 & 3.643 & -3.270 & -1.426 & 1.796 & 1.819
\end{bmatrix}$$

$Y_{\text{train},2} =$

$$[1.631 \quad 1.944 \quad 1.687 \quad 1.496 \quad 1.406 \quad 1.382]^T$$

The estimation values of the test dataset $g(m^2)_{\text{test}}$ of the 10 realizations and the averaged estimation results $Y_{\text{test},2}$ are shown as follows:

$g(m^2)_{\text{test}} = f(m^2, X_{\text{test}}) =$

$$\begin{bmatrix}
3.468 & -0.124 & 1.933 & 2.406 & 4.864 & 3.961 & -1.192 & -0.503 & 1.872 & 1.687 \\
2.715 & -0.856 & 2.173 & 2.851 & 4.768 & 3.728 & -2.312 & -0.926 & 1.807 & 1.795 \\
2.549 & -1.326 & 2.312 & 3.087 & 4.756 & 3.660 & -2.975 & -1.252 & 1.797 & 1.816
\end{bmatrix}$$

$Y_{\text{test},2} =$

$$[1.837 \quad 1.574 \quad 1.442]^T$$

Based on the averaged estimation results and the target values, the loss of training dataset $L_{\text{train},2}$ and the loss of test dataset $L_{\text{test},2}$ at step 2 are calculated according to Eq. (S.1) and shown as follows:

$L_{\text{train},2} = 9.409$

$L_{\text{test},2} = 9.382$

The data mismatch of the model $S_d(m^2)$ is calculated, and its mean $E(S_d(m^2))$ and standard deviation $\sigma(S_d(m^2))$ are shown as follows:

$E(S_d(m^2)) = 194859641$

$\sigma(S_d(m^2)) = 70358432$

Since the mean of the data mismatch decreases, the updated model parameters of this iteration step result in better performance of the neural network, and the updated model parameters are accepted. If the data mismatch increases, the updated model parameters should be rejected, and the latest accepted model parameters from the previous steps are taken as the result of the current iteration step.

*Step 3*

It is shown in the previous subsection that, although the mean of the data mismatch decreases from step 1 to step 2, the standard deviation increases. Thus, the multiplier $\lambda$ should not be changed according to the iteration strategy in Appendix B. The multiplier at step 3 is denoted as $\lambda_2$ since it is determined based on the results at step 2.

$\lambda_2 = \lambda_1 = 17717629$

The perturbed observation matrix $D^2_{obs}$ is generated according to the multivariate Gaussian distribution, which is given as:

$D^2_{obs} =$

$$\begin{bmatrix} 3.002 & 2.996 & 2.999 & 3.002 & 3.004 & 3.004 & 3.008 & 3.000 & 2.995 & 3.004 \\ 4.999 & 4.983 & 4.990 & 4.996 & 4.990 & 4.993 & 5.014 & 4.991 & 5.002 & 4.994 \\ 8.989 & 8.989 & 9.009 & 8.991 & 8.990 & 8.964 & 8.998 & 9.017 & 9.015 & 9.006 \\ 12.953 & 12.996 & 13.017 & 12.998 & 12.990 & 12.999 & 13.027 & 13.016 & 13.047 & 13.023 \\ 16.951 & 16.993 & 17.033 & 16.973 & 17.022 & 17.007 & 16.982 & 16.952 & 16.992 & 16.944 \\ 18.967 & 18.970 & 19.044 & 19.030 & 19.044 & 18.956 & 18.986 & 19.016 & 18.972 & 19.036 \end{bmatrix}$$

The $C_D$ and $C_M$ are defined as prior information, and they do not change among iteration steps. Thus, the $C_D$ and $C_M$ at all of the following steps are the same as the matrices at step 2. Regarding $C_{M_2,D_2}$, $C_{D_2}$ and $C_{M_2}$, they are calculated based on the results at step 2 as follows:

$$C_{M_2,D_2} = \frac{(m^2 - \overline{m^2})(g(m^2)_{train} - \overline{g(m^2)_{train}})^T}{N_e - 1} =$$

$$\begin{bmatrix}
1.604 & 1.428 & 1.776 & 2.028 & 2.166 & 2.199 \\
-0.579 & -0.365 & -0.587 & -0.726 & -0.770 & -0.775 \\
0.269 & 0.220 & 0.159 & 0.152 & 0.136 & 0.126 \\
0.053 & 0.251 & 0.556 & 0.690 & 0.782 & 0.816 \\
0.431 & -0.077 & -0.198 & -0.263 & -0.312 & -0.329 \\
-0.046 & 0.388 & 0.481 & 0.546 & 0.604 & 0.622 \\
-0.078 & -0.536 & -0.994 & -1.219 & -1.330 & -1.365 \\
-0.691 & -0.400 & -0.427 & -0.458 & -0.450 & -0.442 \\
-0.759 & -0.653 & -0.454 & -0.390 & -0.402 & -0.410 \\
-1.852 & -1.133 & -0.820 & -0.727 & -0.698 & -0.692 \\
0.822 & 0.510 & 0.473 & 0.476 & 0.507 & 0.521 \\
-0.169 & -0.079 & -0.326 & -0.466 & -0.534 & -0.549 \\
1.048 & 0.038 & -0.577 & -0.882 & -1.036 & -1.080 \\
-1.015 & -1.130 & -1.159 & -1.243 & -1.298 & -1.309 \\
-0.919 & -0.481 & -0.298 & -0.233 & -0.226 & -0.229 \\
1.604 & 1.204 & 1.056 & 1.033 & 1.051 & 1.058
\end{bmatrix}$$

$$C_{D_2} = \frac{(g(m^2)_{train} - \overline{g(m^2)_{train}})(g(m^2)_{train} - \overline{g(m^2)_{train}})^T}{N_e - 1} =$$

$$\begin{bmatrix}
4.374 & 3.294 & 3.033 & 3.061 & 3.139 & 3.162 \\
3.294 & 3.453 & 3.713 & 3.995 & 4.215 & 4.282 \\
3.033 & 3.713 & 4.453 & 4.980 & 5.331 & 5.436 \\
3.061 & 3.995 & 4.980 & 5.641 & 6.065 & 6.191 \\
3.139 & 4.215 & 5.331 & 6.065 & 6.534 & 6.674 \\
3.162 & 4.282 & 5.436 & 6.191 & 6.674 & 6.818
\end{bmatrix}$$

$$C_{M_2} = \frac{(m^2 - \overline{m^2})(m^2 - \overline{m^2})^T}{N_e - 1} =$$

$$\begin{bmatrix}
1.968 & -1.085 & 0.356 & 0.085 & 0.464 & 0.226 & -0.302 & -0.642 & -0.184 & -0.266 & 0.479 & -1.020 & 0.028 & -0.479 & 0.074 & 0.483 \\
-1.085 & 1.544 & -0.288 & -0.131 & -0.416 & -0.345 & 0.243 & 0.630 & -0.103 & -0.149 & -0.256 & 0.443 & -0.350 & 0.487 & -0.376 & 0.112 \\
0.356 & -0.288 & 0.222 & -0.251 & 0.088 & 0.101 & 0.072 & -0.194 & -0.050 & -0.094 & -0.024 & -0.121 & 0.099 & -0.263 & 0.106 & 0.040 \\
0.085 & -0.131 & -0.251 & 0.635 & 0.026 & 0.052 & -0.445 & 0.013 & 0.069 & 0.237 & 0.199 & -0.038 & -0.122 & 0.196 & 0.024 & 0.033 \\
0.464 & -0.416 & 0.088 & 0.026 & 0.450 & -0.216 & 0.006 & -0.377 & 0.174 & -0.004 & 0.098 & -0.329 & 0.428 & 0.105 & 0.095 & 0.158 \\
0.226 & -0.345 & 0.101 & 0.052 & -0.216 & 0.918 & 0.022 & 0.152 & -0.541 & 0.221 & 0.335 & 0.156 & 0.008 & -0.271 & 0.113 & -0.081 \\
-0.302 & 0.243 & 0.072 & -0.445 & 0.006 & 0.022 & 0.970 & 0.351 & -0.458 & -0.103 & 0.093 & -0.110 & 0.532 & -0.130 & -0.171 & -0.082 \\
-0.642 & 0.630 & -0.194 & 0.013 & -0.377 & 0.152 & 0.351 & 0.599 & -0.341 & 0.125 & 0.028 & 0.182 & -0.198 & 0.120 & -0.168 & -0.220 \\
-0.184 & -0.103 & -0.050 & 0.069 & 0.174 & -0.541 & -0.458 & -0.341 & 0.981 & 0.571 & -0.575 & -0.088 & -0.432 & 0.328 & 0.450 & -0.243 \\
-0.266 & -0.149 & -0.094 & 0.237 & -0.004 & 0.221 & -0.103 & 0.125 & 0.571 & 1.575 & -0.425 & -0.311 & -0.550 & 0.351 & 0.896 & -0.681 \\
0.479 & -0.256 & -0.024 & 0.199 & 0.098 & 0.335 & 0.093 & 0.028 & -0.575 & -0.425 & 0.610 & -0.104 & 0.411 & -0.049 & -0.361 & 0.282 \\
-1.020 & 0.443 & -0.121 & -0.038 & -0.329 & 0.156 & -0.110 & 0.182 & -0.088 & -0.311 & -0.104 & 1.014 & 0.162 & 0.049 & -0.222 & 0.005 \\
0.028 & -0.350 & 0.099 & -0.122 & 0.428 & 0.008 & 0.532 & -0.198 & -0.432 & -0.550 & 0.411 & 0.162 & 1.283 & -0.122 & -0.276 & 0.332 \\
-0.479 & 0.487 & -0.263 & 0.196 & 0.105 & -0.271 & -0.130 & 0.120 & 0.328 & 0.351 & -0.049 & 0.049 & -0.122 & 0.877 & -0.001 & -0.154 \\
0.074 & -0.376 & 0.106 & 0.024 & 0.095 & 0.113 & -0.171 & -0.168 & 0.450 & 0.896 & -0.361 & -0.222 & -0.276 & -0.001 & 0.692 & -0.398 \\
0.483 & 0.112 & 0.040 & 0.033 & 0.158 & -0.081 & -0.082 & -0.220 & -0.243 & -0.681 & 0.282 & 0.005 & 0.332 & -0.154 & -0.398 & 0.726
\end{bmatrix}$$

The model parameters at step 3 can be updated based on the model parameters from step 2 ($m^2$), the estimation results ($g(m^2)_{train}$), the step size ($\lambda_2$), the perturbed observation ($D^2_{obs}$), the cross-covariance matrix ($C_{M_2,D_2}$), and the covariance matrices ($C_{D_2}$, and $C_{M_2}$) according to Eq. (10).

The ultimate result of the model parameters at step 3 is shown as follows:

$m^3 =$

$$\begin{bmatrix}
1.430 & 2.014 & 1.732 & 0.825 & 1.473 & 2.016 & -1.199 & 0.324 & -0.524 & 2.953 \\
1.030 & -0.540 & 0.483 & 0.261 & 0.481 & -1.648 & 2.328 & -1.395 & 1.146 & -0.659 \\
-0.563 & 0.322 & -1.007 & -1.124 & -0.430 & -0.440 & -1.330 & -0.706 & -0.450 & -0.366 \\
0.284 & -0.810 & 1.830 & 1.241 & 0.435 & 1.695 & 0.999 & 0.875 & 0.261 & 0.722 \\
-0.705 & 0.511 & -0.223 & -0.614 & -0.392 & 0.518 & -0.215 & -0.418 & -1.782 & 0.058 \\
1.873 & 0.040 & 0.315 & 0.197 & -0.158 & 1.730 & -0.011 & 1.328 & 2.199 & 1.906 \\
0.576 & -0.320 & -1.872 & -1.661 & -2.310 & -2.182 & -1.287 & -0.902 & -1.763 & -2.112 \\
0.687 & -1.291 & -0.005 & -0.236 & -1.186 & -1.523 & -0.015 & -0.269 & 0.388 & -0.994 \\
-2.792 & 0.301 & -0.528 & 0.962 & -0.341 & -0.244 & -0.040 & -0.935 & -0.618 & -0.555 \\
-2.676 & -1.214 & -0.880 & -0.289 & -4.261 & -0.645 & -1.689 & -2.144 & -0.568 & -1.404 \\
1.457 & -0.720 & 0.432 & -0.503 & 0.201 & 0.576 & 0.014 & 0.842 & -0.709 & 1.111 \\
-0.314 & -1.256 & -1.393 & -0.687 & 0.724 & 0.116 & 1.222 & 0.634 & 1.366 & -0.770 \\
0.465 & -0.122 & -1.466 & -2.169 & -0.588 & 0.677 & 0.055 & 0.425 & -2.219 & -1.180 \\
-2.074 & -1.852 & -0.653 & -1.085 & -1.825 & -1.756 & 0.506 & -1.608 & -1.757 & -0.221 \\
-2.101 & 0.005 & -0.633 & -0.586 & -2.143 & 0.060 & -1.503 & -1.304 & -0.153 & -0.943 \\
2.130 & 0.998 & 0.682 & 0.944 & 2.404 & 1.776 & 1.704 & 0.195 & 0.283 & 1.401
\end{bmatrix}$$

Based on the updated model parameters $m^3$, and the inputs of the training data $X_{train}$, the estimations of the training dataset ($g(m^3)_{train}$ and $Y_{train,3}$) are obtained as follows:

$g(m^3)_{train} = f(m^3, X_{train}) =$

$$\begin{bmatrix}
4.368 & 2.564 & 1.211 & 2.012 & 6.860 & 4.451 & 2.288 & 2.567 & 0.434 & 2.217 \\
4.132 & 2.164 & 2.817 & 3.207 & 6.716 & 4.800 & 1.461 & 2.406 & 3.285 & 2.531 \\
3.582 & 1.432 & 4.519 & 3.749 & 6.505 & 4.722 & 1.354 & 2.273 & 3.689 & 3.011 \\
3.470 & 0.990 & 4.922 & 3.792 & 6.450 & 4.642 & 1.355 & 2.543 & 3.749 & 3.214 \\
3.462 & 0.826 & 5.112 & 3.795 & 6.440 & 4.616 & 1.360 & 2.764 & 3.755 & 3.267 \\
3.462 & 0.793 & 5.163 & 3.796 & 6.439 & 4.612 & 1.363 & 2.820 & 3.755 & 3.274
\end{bmatrix}$$

$Y_{train,3} =$

$$\begin{bmatrix} 2.897 & 3.352 & 3.484 & 3.513 & 3.540 & 3.548 \end{bmatrix}^T$$

The estimation values of the test dataset $g(m^3)_{test}$ of the 10 realizations at step 3 and the averaged estimation results $Y_{test,3}$ are shown as follows:

$g(m^3)_{test} = f(m^3, X_{test}) =$

$$\begin{bmatrix}
3.814 & 1.782 & 4.046 & 3.631 & 6.583 & 4.772 & 1.363 & 2.257 & 3.598 & 2.804 \\
3.494 & 1.166 & 4.761 & 3.782 & 6.467 & 4.674 & 1.354 & 2.392 & 3.734 & 3.142 \\
3.464 & 0.885 & 5.035 & 3.795 & 6.443 & 4.624 & 1.357 & 2.673 & 3.754 & 3.250
\end{bmatrix}$$

$Y_{test,3} =$

$$\begin{bmatrix} 3.465 & 3.497 & 3.528 \end{bmatrix}^T$$

The loss of the training dataset $L_{train,3}$ and the loss of the test dataset $L_{test,3}$ at step 3 are obtained based on the averaged estimation results and the target values.

$L_{\text{train},3} = 7.611$
$L_{\text{test},3} = 7.503$

The data mismatch of the model $S_d(m^3)$ is calculated, and its mean and the standard deviation are shown as follows:

$E(S_d(m^3)) = 139406121$
$\sigma(S_d(m^3)) = 35855126$

Since both the mean and the standard deviation of the data mismatch decreases, the updated model parameters are accepted and $\lambda$ is reduced to increase the step size at step 4.

*Step 4 to step 31*

The detailed calculation process from step 4 to step 31 is not shown due to the limited space of the article. However, the estimation value, the estimation loss, and the $\lambda$ (step size) of the iteration steps are summarized and compared in figures to demonstrate the converging process. The estimation values of the training dataset and the test dataset are shown in **Fig. S.1**. All of the 10 realizations of each training data point are denoted in black lines. The realizations of the three test data points are represented in green line, red line, and blue line, respectively. The target values of both training dataset and test dataset are represented in red pentacles. It is shown in **Fig. S.1** that the estimation values of the 10 realizations of each data point converge to the target values as the iteration step increases.

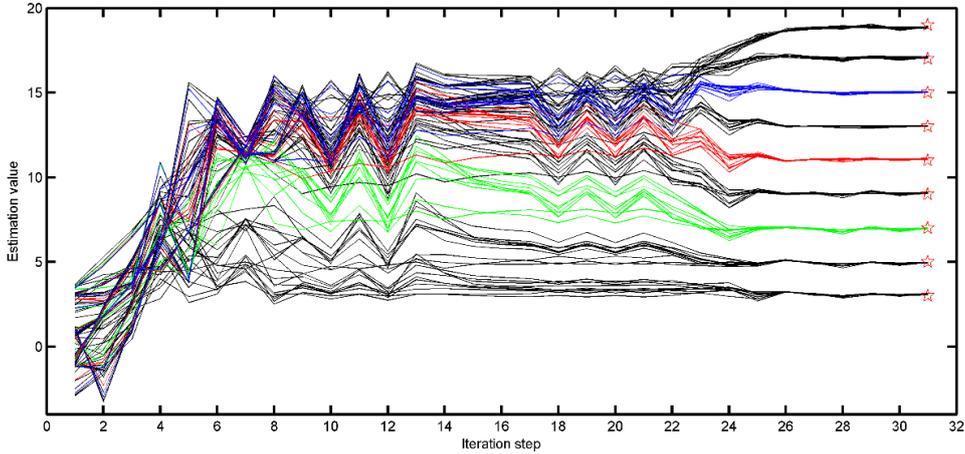

**Fig. S.1.** Estimation values of the training dataset (black lines) and the test dataset (blue, red, and green lines). Each data point has 10 realizations, and the estimation values converge to the target value (red pentacles) as iteration step increases.

Regarding the estimation loss, the red line in **Fig. S.2** represents the estimation loss of the training dataset, and the blue line denotes the estimation loss of the test dataset. Both the estimation losses of the training dataset and the test dataset converge to 0 as the training progresses. In addition, the $\lambda$ of different iteration steps are shown in **Fig. S.3**. The $\lambda$ is inversely proportional to the step size, and it is updated according to the Appendix B. The decrease of $\lambda$ between step 22 and step 27 indicates that both the mean and the standard deviation of the data mismatch are decreasing, which is consistent with the trends in **Fig. S.1** and **Fig. S.2**. This implies that the model is in the right searching direction, and larger step size is used to speed up the converging process. The increase of $\lambda$ between step 28 and step 31 means that we cannot find a better result with smaller data mismatch, which indicates that the result is close to the local or global minimum.

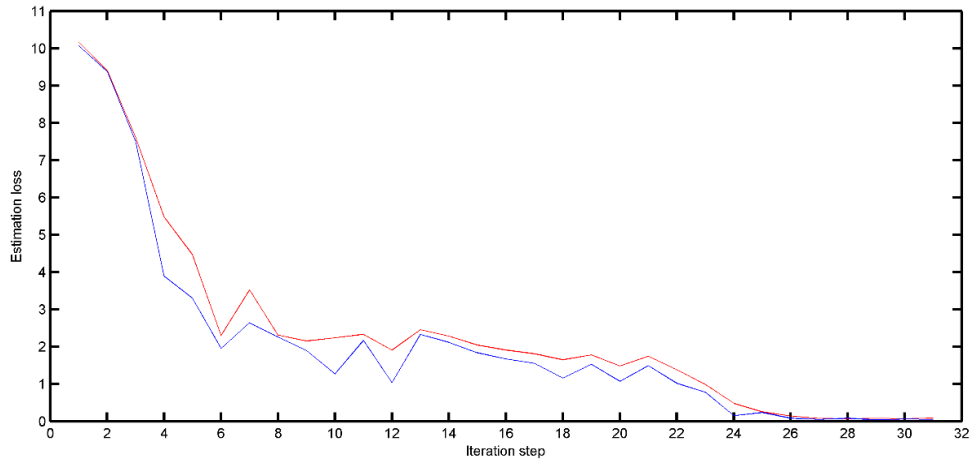

**Fig. S.2.** Estimation losses of the training dataset (red line) and the test dataset (blue line).

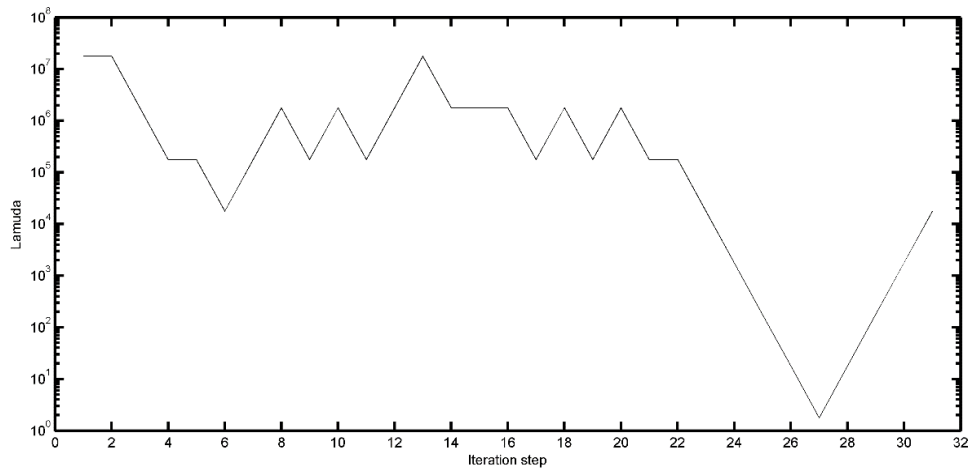

**Fig. S.3.** Value of $\lambda$ at different iteration steps. The smaller the $\lambda$, the larger the step size. Larger step sizes are applied when we have confidence in the update direction. Smaller step sizes are used when we cannot find better results, which means that the result is close to the local or global minimum.

The converging process of the model parameters is illustrated in **Fig. S.4**. Each subgraph shows a model parameter matrix at an iteration step with its columns denoting the 10 realizations and its rows representing different weights, and the values of the model parameters are shown in colors via a grey-scale map. It is shown in **Fig. S.4** that the columns of the subgraph at step 1 are totally different from each other since the initial model parameters are generated from a multivariate Gaussian distribution. Compared with the subgraph at step 3, the subgraph at step 7 has very similar columns, which indicates that the realizations are converging at a high speed from step 3 to step 7.

This is consistent with the rapid drop of estimation loss in **Fig. S.2**. The subgraph at step 27 is similar to that at step 31, which implies that there are few updates from step 27 to step 31 and the model has found the local or global minimum. This is also proven by the trend of the estimation values in **Fig. S.1** and the estimation loss in **Fig. S.2**.

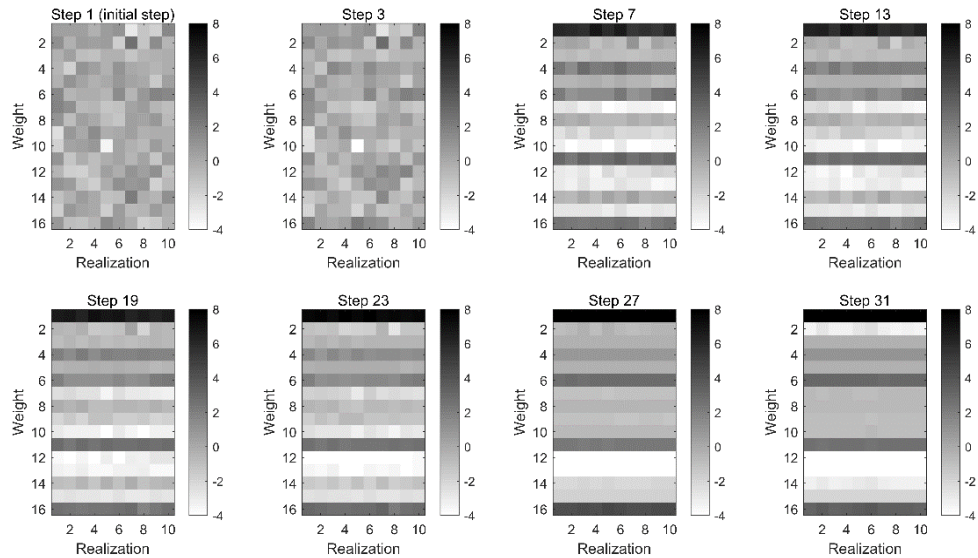

**Fig. S.4.** Converging process of the model parameters at different iteration steps. The values of model parameters are represented via a grey-scale map. Each column represents the 16 weights of a certain realization. Each row denotes the same weight in the 10 realizations.